\documentclass[twoside,leqno,twocolumn]{article}

\usepackage[letterpaper]{geometry}

\usepackage{ltexpprt}
\usepackage[utf8]{inputenc} 
\usepackage[T1]{fontenc}    
\usepackage{hyperref}       
\usepackage{url}            
\usepackage{booktabs}       
\usepackage{amsfonts}       
\usepackage{nicefrac}       
\usepackage{microtype}      
\usepackage{amsmath}
\usepackage{bbm}
\usepackage{color,soul}
\usepackage{graphicx}
\usepackage{multirow}
\usepackage{subcaption}
\usepackage{wrapfig}
\usepackage{bm}

\DeclareMathOperator*{\argmax}{arg\,max}
\newcommand{\blue}[1]{{\color{blue} \hl{[#1]}}}

\begin{document}

\title{HALO: Learning to Prune Neural Networks with Shrinkage}
\author{%
 Skyler Seto, Martin T. Wells and Wenyu Zhang\\
  Department of Statistics and Data Science\\
  Cornell University\\
  Ithaca, NY \\
  \texttt{$\{$ss3349, mtw1, wz258$\}$@cornell.edu} \\}

\date{}

\maketitle


\fancyfoot[R]{\scriptsize{Copyright \textcopyright\ 2021 by SIAM\\
Unauthorized reproduction of this article is prohibited}}





\begin{abstract}
    Deep neural networks achieve state-of-the-art performance in a variety of tasks by extracting a rich set of features from unstructured data, however this performance is closely tied to model size. Modern techniques for inducing sparsity and reducing model size are (1) network pruning, (2) training with a sparsity inducing penalty, and (3) training a binary mask jointly with the weights of the network.  We study these approaches from the perspective of Bayesian hierarchical models and present a novel penalty called \textbf{H}ierarchical \textbf{A}daptive \textbf{L}ass\textbf{o} (HALO) which learns to sparsify weights of a given network via trainable parameters.  When used to train over-parametrized networks, our penalty yields small subnetworks with high accuracy without fine-tuning. Empirically, on image recognition tasks, we find that HALO is able to learn highly sparse network (only $5\%$ of the parameters) with significant gains in performance over state-of-the-art magnitude pruning methods at the same level of sparsity. Code\footnote{\url{https://github.com/skyler120/sparsity-halo}} is available at the provided link.
\end{abstract}


\section{Introduction}

Machine learning systems have improved many modern technologies including web search systems, recommendation systems, and cameras. Traditional machine learning systems rely on human experts to extract features from the raw data in order to perform classification. As such, these systems are limited by the features that humans design. Representation learning in contrast automatically discovers relevant features from the raw data. 

Deep neural networks (DNNs) are representation learning methods that learn representations of data by composing non-linear functions that transform the input through a series of compositions. With enough compositions, these deep learning systems can model arbitrarily complex functions. For example, a convolutional neural network operating on images learns features by taking the raw image and applies a series of convolutions over patches of the image to generate spatial features. In a convolutional neural network, the learned features in the first layer may be edge detectors, in the second layer may detect arrangements of edges, and in the third layer recognize basic objects \cite{lecun2015deep}.  As the deep neural network is trained for a specific task (such as image classification) the features obtained in the later layers of the network become more specialized for the task.

\begin{figure}[t]
    \centering
    \includegraphics[width=\linewidth]{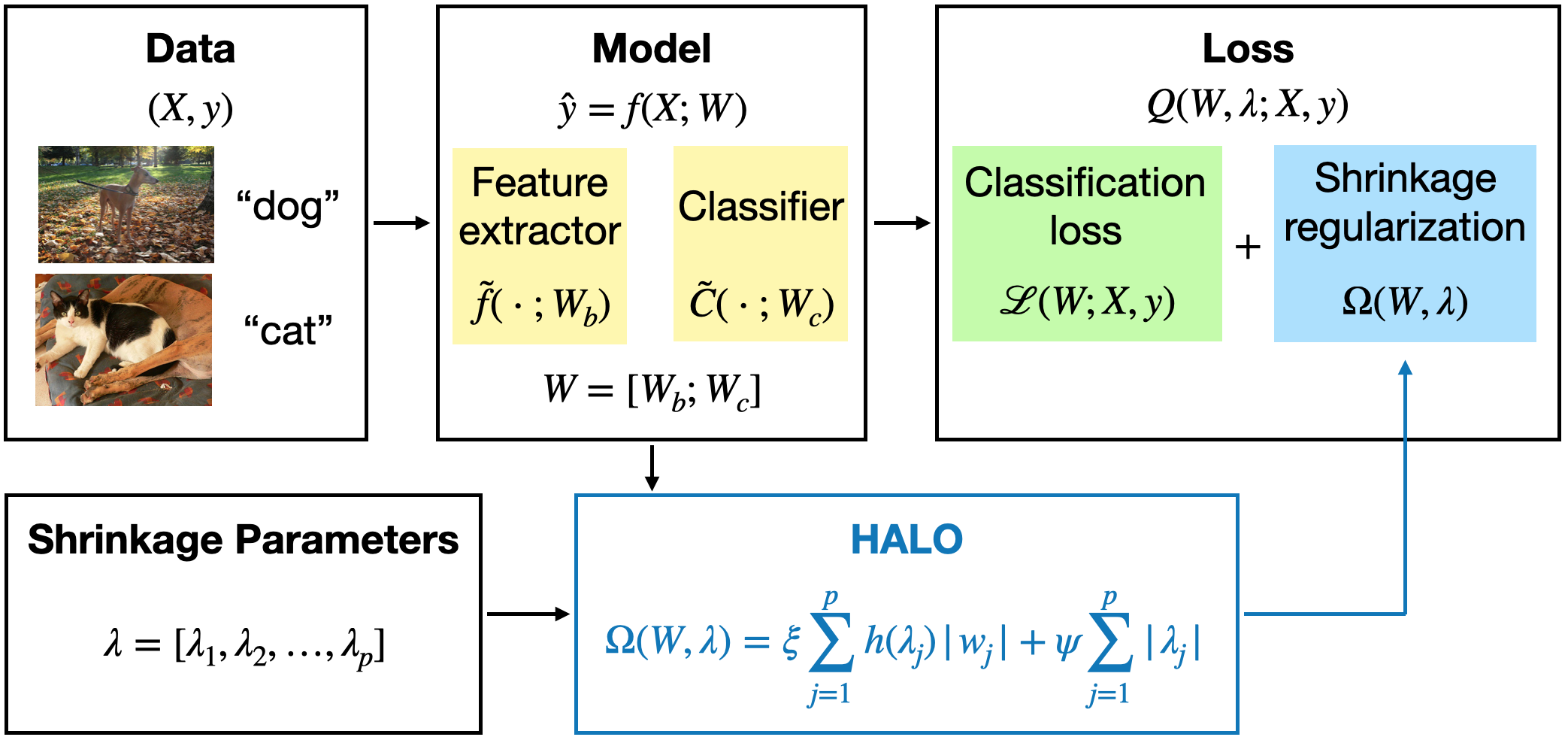}
    \caption{Overview of HALO: we perform MAP estimation of our Bayesian hierarchical model by appending an additional penalty to the standard loss function which enforces sparsity via shrinkage on the parameters of the network.}
    \label{fig:halo}
\end{figure}

Due to their ability to extract complex feature representations, neural networks have achieved state-of-the-art performance on numerous problems in image recognition \cite{krizhevsky2012imagenet}, speech recognition \cite{hinton2012deep}, natural language understanding \cite{devlin2018bert}, and healthcare \cite{esteva2019guide}.  However, this performance is closely tied to model size, since DNNs rely on compositions of numerous non-linear functions to extract features and will often contain millions of parameters. In settings where memory footprint and computational efficiency are important such as on low-power devices, DNNs although favored over smaller models with limited performance are unable to be deployed. 

In this work, we propose a method to learn sparse neural networks that match the performance of over-parametrized networks with little performance reduction.  We motivate our approach as a Bayesian hierarchical model which adaptively shrinks the  weights of a network. We derive the MAP estimate for this model and call this penalty \textbf{H}ierarchical \textbf{A}daptive \textbf{L}ass\textbf{o} (HALO). HALO regularizes model parameters in a hierarchical fashion and shrinks each model parameter based on its importance to the data and model. With the resulting order in the magnitude of the model parameters, further pruning by simple thresholding can be applied to obtain the desired level of sparsity with less drop in accuracy than competing methods. We demonstrate on multiple image recognition tasks and neural network architectures that HALO is able to learn highly sparse feature extractors with little to no accuracy drop. An overview of the additions we make to the standard training pipeline is provided in Figure~\ref{fig:halo}.

\section{Related Work}
\label{sec:rw}

\subsection{Pruning Methods}

are a technique for compressing neural networks which first used a criteria based on the Hessian of the loss function \cite{hassibi1993optimal, lecun1990optimal} to remove weights. Recent work also explored pruning individual weights based on their overall contribution to the loss \cite{lee2018snip, wang2020picking} and the magnitude based criteria $|w_j|_1$ where $w_j$ is a weight parameter of the network \cite{guo2016dynamic, han2015deep, narang2017exploring}. These networks are initialized based on weights from the first iteration requiring training until full convergence of a full model.

Recent work referred to as the lottery ticket hypothesis (LTH) \cite{frankle2018lottery, zhou2019deconstructing} explored the magnitude criteria and demonstrated small subnetworks are trainable from scratch. Their approach was to first train a randomly initialized over-parameterized network, threshold small weights to zero, then re-initialize the non-zero weights to the same initialization as the over-parameterized network and train only the non-zero weights. This procedure determined that the individual weights are important and not the final weight values from the first stage of training. 
Other recent works expanded on the LTH by identifying subnetworks in a randomly initialized network which perform well on a given task without training the weights \cite{ramanujan2020s}. Specifically, Ramanujan et al. \cite{ramanujan2020s} devised an algorithm for finding randomly initialized subnetworks in larger over-parameterized networks that performed better than trained networks, and Malch et al. \cite{malach2020proving} proved that for any network of depth $\ell$, a subnetwork could be found in any depth $2\ell$ network which achieves equivalent performance to the depth $\ell$ network.  

\subsection{Learning Masks and Weights} 

is a one-stage procedure for learning sparse neural networks through learning a binary mask over the network's parameters. This pruning problem is formulated as 

\vspace{-5mm}
\begin{equation}
    Q(W, M | X, y) = \mathcal{L}(h(M)\odot W | X, y) + \xi \Omega(M),
    \label{eq:mask}
\end{equation}
\vspace{-5mm}

\noindent where $\Omega(\cdot)$ is a penalty function according to some pre-specified criteria, $\mathcal{L}$ is the standard loss function e.g. cross entropy for classification,  $W=[w_1, w_2 \dots, w_p]$ are the model parameters, $M$ are additional trainable parameters, $h(M)$ is a pre-specified mask function $h: \mathbb{R} \rightarrow \{0, 1\}$, $\xi$ is a non-negative hyperparameter controlling the trade-off between the loss and penalty, and in some cases additional penalties might be applied to $W$. Numerous works suggest different approaches for selecting the mask function $h$ \cite{azarian2020learned, kusupati2020soft, liu2020dynamic, savarese2019winning, wang2020picking, xiao2019autoprune}. Other works explore a Bayesian model for the mask formulation \cite{kondo2016bayesian, louizos2017learning} which we explore in Section~\ref{sec:method} where we discuss point mass priors for inducing sparsity in Bayesian hierarchical models. While this class of models has optimal frequentist properties, the posterior is computationally intractable in many cases and difficult to optimize, and further does not perform any feature selection of non-zero weights. Although \eqref{eq:mask} appears to be very similar to our objective, we note that the class of estimators our approach is based on and this approach are quite different in how they prune models. We elaborate on this difference in Section~\ref{sec:bhp}.

\subsection{Regularization Methods}

induce sparsity in deep neural networks using a pre-specified criteria (usually referred to as a penalty). In the regularization setting, we consider the modified loss

\vspace{-6mm}
\begin{equation}
    \label{eq:objective}
Q(W|X, y) = \mathcal{L}(W | X, y) + \xi\Omega(W),
\end{equation}
\vspace{-6mm}

Early works on sparsity in neural networks augment the loss function with a penalty to attain sparse neural networks and limit overfitting  \cite{carreira2018learning, chauvin1989back, ishikawa1996structural, weigend1991generalization}.  The $L_0$ penalty induces sparse models by penalizing the number of non-zero entries in $W$ without any further bias on the weights $W$ of the model. A problem is that the penalty is computationally intractable as it is non-differentiable and the learning problem is NP-hard. An alternative to $L_0$ regularization is $L_1$ regularization obtained by adding a penalty on the magnitude of the weights, its tightest convex relaxation.  The associated estimator is called the Lasso estimator \cite{tibshirani1996regression}.  Although the Lasso has strong oracle properties under certain conditions, it is a biased estimator \cite{fan2001variable}.  The Lasso requires a neighborhood stability/strong irrepresentable condition on the design matrix $X$ for the selection consistency \cite{tropp2006just, wainwright2009sharp, zhao2006model}.

 More recently, Collins et al. \cite{collins2014memory} applied $L_q$ norm penalties (also called bridge penalties) to achieve sparse networks with $4X$ memory compression over the original network and a minimal decrease in accuracy on ImageNet \cite{krizhevsky2012imagenet}. The standard $L_q$ norm penalty is written as $\Omega_q(W) = \left(\sum_{j=1}^p |w_j|^q)\right)^{1/q}$ for which the Lasso $q=1$ and ridge (weight decay) $q=2$ penalties are special cases. 
Other extensions include nonconvex penalties like minimax concave penalty (MCP) \cite{zhang2010nearly} and smoothly clipped absolute deviations (SCAD) \cite{fan2001variable} have also been applied to deep neural networks \cite{vettam2019regularized}. Extensions of the Lasso aimed at correcting the bias in the Lasso estimator including  the trimmed Lasso \cite{yun2019trimming} and other nonconvex penalties like minimax concave penalty (MCP) \cite{zhang2010nearly} and smoothly clipped absolute deviations (SCAD) \cite{fan2001variable} have also been applied to deep neural networks \cite{vettam2019regularized}.  However these studies were limited to smaller image classification datasets such as MNIST and Fashion-MNIST. They found marginal improvements over the $L_1$ penalty in some cases, and did not aim for highly sparse networks. Recently \cite{pieper2020nonconvex, Taheri2020} examined theoretical properties of Lasso-type and non-convex regularization for neural networks.  

Other works also induce sparsity through Bayesian hierarchical models \cite{molchanov2017variational, nalisnick2018dropout, polson2018posterior, ullrich2017soft}; similarly in Section~\ref{sec:halo} we discuss sparsity and posterior convergence of the Bayesian hierarchical model corresponding to HALO.

\textbf{Algorithms for Optimizing Regularizers:} Contemporary literature has explored algorithms optimizing regularizers for better generalization performance \cite{lorraine2018stochastic, nakamura2019adaptive, streeter2019learning}; the latter of which optimizes the weight decay parameter, which is similar to our approach, however weight decay does not directly induce sparsity as heavily as $L_1$ norm penalties.  Further, although we optimize our regularization coefficients jointly with the training data, they could be optimized over a validation set for improving generalization and test set performance.

\section{Sparse Penalties and Hierarchical Priors}
In this section, we distinguish between the sparse penalty approach (HALO approach), magnitude pruning strategies and binary masking through the lens of Bayesian shrinkage estimation. 

\subsection{Sparse Penalties}
\label{sec:method}

\subsubsection{The Weighted Lasso and Pruning}
One approach to reducing the bias in Lasso is to select a different regularization coefficient for each parameter resulting in the weighted Lasso :

\vspace{-6mm}
\begin{equation}
    \label{eq:wlasso}
    \Omega_{\text{weighted}}(W) = \sum_{j=1}^p \lambda_j|w_j|,
\end{equation}
\vspace{-4mm}

\noindent or adaptive Lasso penalty\cite{zou2006adaptive} which sets $\lambda_j = \frac{1}{\hat{w}_j}$ where $\hat{w}_j$ is an initial estimate from another run of OLS or Lasso. With enough training data (oracle property) and relatively lower noise, adaptive lasso could often select variables accurately. However, it is computationally intensive and does not work well with extreme correlation \cite{zhang2014comparisons}.

A more general form for the adaptive Lasso which extends to other sparsity-inducing penalties and any general loss function $\mathcal{L}(\cdot)$ is known as the local linear approximation algorithm (LLA) which iteratively solves the objective in $k$ iterations:

\vspace{-6mm}
\begin{equation}
    \label{eq:lla}
    W^{(k+1)} = \argmax_W \left[ \mathcal{L}(W) - \sum_{j=1}^p\Omega(w_j^{(k)})|w_j| \right]
\end{equation}
\vspace{-4mm}

\noindent where $w_j^{(k)}$ are the weights for the previous iteration of LLA \cite{zou2008one}. Note that by running this optimization twice with $\Omega_1$ and thresholding the coefficients based on the first run, we recover magnitude pruning \cite{han2015deep} and can view magnitude pruning as an adaptive Lasso estimator.

\subsubsection{Nonconvex Penalties}
An alternative approach is to use a penalty that diminishes in value for large parameter values. These types of penalties are non-convex but  yield both empirical and theoretical results \cite{vettam2019regularized, zhang2010nearly, zou2008one}. Fan and Li \cite{ fan2001variable} proposed a non-convex penalty, smoothly clipped absolute deviation (SCAD) penalty,  to remove the bias of the Lasso and proved an oracle property for one of the local minimizers of the resulting penalized loss.  Zhang \cite{zhang2010nearly} proposed another non-convex penalization approach, minimax concave penalty (MCP):

\vspace{-7.5mm}
\begin{equation}
    \Omega_{MCP}(W; \gamma, \lambda) = \begin{cases}\lambda|w_j| - \frac{w_j^2}{\gamma} & |w_j| \leq \gamma\lambda\\
    \frac{\gamma\lambda^2}{2} & else
    \end{cases}
    \label{eq:mcp}
\end{equation}
\vspace{-4mm}

\noindent where $\gamma$ and $\lambda$ are hyperparameters of the MCP. 

It has been shown that for some nonconvex penalty functions such as the SCAD penalty or MCP that the LLA yields an optimal solution when $k=1$, and as such nonconvex penalties are a more efficient class of penalties \cite{strawderman2013hierarchical, zou2008one}. Further, this class of nonconvex penalties is preferred to other penalties as they are shown to be the optimal class of penalties for achieving sparsity and unbiasedness of the regression parameter estimates \cite{zou2008one}.

\subsection{Bayesian Hierarhical Priors}
\label{sec:bhp}

Bayesian hierarchical models are useful modeling and estimation approaches since their model structure allows “borrowing strength” in estimation. This means that the prior affects the posterior distribution by shrinking the estimates towards a central value. 
From a Bayesian point of view we can consider (\ref {eq:objective}) as a log posterior density, and with this interpretation the penalty $ \xi\Omega(W)$ can then be identified with a log prior distribution of $W$.  Constructing estimates via optimization of (\ref {eq:objective}) then gives a maximum a posteriori (MAP) estimation procedure. 

The Lasso consists of a Laplace($\lambda$) prior on model parameters $W$. Strawderman et al. \cite{strawderman2013hierarchical} study the estimator given by \eqref{eq:mcp} from a hierarchical Bayes perspective. The intuition  is that $\exp\{\Omega_{MCP}(W, \lambda; \gamma)\}$ is a hierarchical prior with the first level being a Laplace ($\lambda$) prior on $W$ (as with the Bayesian Lasso) and the second level is a half normal prior on the hyperparameter $\lambda$.  
They further studied the priors of the corresponding hierarchical Bayes procedure as part of a class of scale mixture priors similar to those used in dropout \cite{nalisnick2018dropout} and demonstrated that the MAP estimate for this procedure is equivalent to optimization with the MCP penalty for the linear model \cite[Remark~4.3]{strawderman2013hierarchical}.  In our experiments we denote the MAP estimate of MCP as the SWS penalty. 


An alternative is a mixture of a point mass at zero and a continuous distribution $g$,

\vspace{-7mm}
\begin{equation}
    p(w) = \zeta g(w) + (1 - \zeta) \delta_0, \hspace{5mm} p(\zeta) = \text{Bern}(\pi)
\end{equation}
\vspace{-7mm}

\noindent which are referred to as spike and slab priors \cite{mitchell1988bayesian} .  
Variants of this formulation have been studied recently for pruning deep neural networks where a MAP estimate is approximated by learning a continuous function representing a mask over the weights of the network \cite{azarian2020learned, kusupati2020soft, liu2020dynamic, louizos2017learning, savarese2019winning, wang2020picking, xiao2019autoprune}.

A primary drawback of spike and slab priors is that computation is much more demanding than for single component continuous shrinkage priors ~\cite{piironen2017sparsity} since sampling of the point mass part of the posterior distribution entails searching over an enormous set of binary indicators and is not feasible in even moderately large parameter spaces. Additionally this class of priors may not effectively penalize the non-zero parameters in $W$ leading to worse predictive performance over normal scale mixture priors.

\section{The Hierarchical Adaptive Lasso (HALO)}
\label{sec:halo}

Although the MCP has desirable properties among shrinkage estimators, a primary drawback is that all weights $w \geq \gamma\lambda$ of the model are penalized equally which can result in over-sparsification. For the standard MCP $\gamma$ and $\lambda$ are both hyperparameters of the model which must be provided apriori and directly influence the sparisty of the model, and in the hierarchical model, they are derived from the hierarchical Gamma and truncated normal priors \cite{strawderman2013hierarchical}. We extend this model by considering an additional level of hierarchy and by making the penalty adaptive such that each $w_j$ has its own $\lambda_j$ in the scale mixture of normals representation. We consider a first level Laplace prior and further place a mixing distribution on the $\lambda_j 's$.  Specifically we define the hierarchical  prior as 

\vspace{-6mm}
\begin{align}
    p(y | \sigma) &= \mathcal{N}(f(X; W), \sigma^2I_p) \label{eq:bayes_halo1}\\
    p(w_j | \alpha_j, \sigma)  &=   Laplace\left(\frac{1}{\xi\alpha_j\sigma^2}\right)\\
    p(\lambda_j | \eta, \sigma, \psi)  &=  Gamma\left(\eta+1, \frac{\psi}{\sigma^2}\right)
    \label{eq:bayes_halo}
\end{align}
\vspace{-5mm}

\noindent where $\alpha_j = 1/\lambda_j^\eta$ and $\eta+1>0$.
Note that the Laplace distribution can be viewed as a scale mixture of normals with an exponential mixing distribution \cite{armagan}.
The second level Gamma prior mixing distribution on the natural parameter of the exponential is related to the class of the exponential-gamma prior distributions developed in \cite{bernardo2009bayesian, griffin2005alternative}. This additional level in the hierarchy is similar to the Horseshoe+ prior which consists of two positive Cauchy distributions \cite{bhadra2017horseshoe+}, and is in contrast to the single level mixture of normals prior used for dropout in Table 1 of \cite{nalisnick2018dropout}.  The additional level of the hierarchy in (\ref{eq:adph_reg}) allows for additional shrinkage and sparsity over the simpler penalties such as SWS.

Estimation and computation of the posterior distribution for this model can be difficult especially for neural networks with millions of parameters. Instead, by denoting $\alpha_j = h(\lambda_j)$, we obtain a generalized MAP estimate for this model:

\vspace{-7mm}
\begin{equation}
    \label{eq:adph_reg}
    \Omega(W, \lambda; \psi, \xi) = \xi\sum_{j=1}^p h(\lambda_j) |w_j| + \psi \sum_{j=1}^p|\lambda_j|.
\end{equation}
\vspace{-5mm}

\noindent where $h(\cdot)$ is a positive function for a generalized version of the penalty.  We call \eqref{eq:adph_reg} the Hierarchical Adaptive Lasso (HALO) penalty since the hierarchical and adaptive penalty places an additional $L_1$ norm on the regularization coefficients $\lambda_j$. We will call the prior defined by \eqref{eq:bayes_halo1} - \eqref{eq:bayes_halo} the HALO prior.  For the MAP estimate both $\lambda$ and $W$ are trainable parameters in the optimization allowing for learning of the appropriate amount of shrinkage and the weights of the model. 

In our experiments, we set $h(\lambda) = 1/\lambda^2$ so that $h(x) \rightarrow \infty$ as $x \rightarrow 0$; this combined with the $L_1$ penalty on $\lambda_j$ encourages selective shrinkage of the weights where important weights remain unregularized, and makes HALO a monotonic penalty \cite{bogdan2013statistical, feng2019sorted}.  This is more flexible than adaptive Lasso methods which fix regularization coefficients each iteration. We modify the SWS penalty to have the same functional penalty as well, and in the supplementary material , we explore and suggest alternatives for $h(\cdot)$. Additionally the theorem below gives conditions under which using HALO in \eqref{eq:objective} is convex. A proof of Theorem 1 is provided in the supplementary material .


\begin{theorem}
    Consider the objective for the penalized linear model
        \vspace{-3mm}
        \begin{align*}
        L(W, \bm{\lambda}) &= L(W) + \Omega(W, \bm{\lambda}; \psi, \xi) \\
        &= \|y - \sum_{j=1}^p x_jw_j\|^2 + \xi\sum_{j=1}^p \frac{1}{\lambda_j^2} |w_j| + \psi \sum_{j=1}^p|\lambda_j|,
        \end{align*}
        
        \noindent and let $X$ be a full rank $n \times p$ matrix with smallest singular value $\nu$. Define 
        \vspace{-3mm}
        $$\mathbbm{R}_{\lambda, W} = \left\{W, \lambda: 24\nu \frac{\lambda_p^{10}}{|w_p|} \leq \xi^3 \leq 24\nu \frac{\lambda_1^{10}}{|w_1|}\right\}$$
        with $\frac{\lambda_1^{10}}{|w_1|} \geq \frac{\lambda_2^{10}}{|w_2|} \geq \dots \geq \frac{\lambda_p^{10}}{|w_p|}$. Then $\nabla^2 L(W, \bm{\lambda}) \succ 0$ and $L(W, \bm{\lambda})$ is elementwise convex over $\mathbbm{R}_{\lambda, W}$. 
\end{theorem}

\subsection{Posterior Concentration}

We will next give a theoretical development for the penalty in (\ref{eq:adph_reg}). An assessment of the goodness of an estimator is some measure of center of the posterior distribution, such as the posterior mean or mode. The natural object to use for assessing feature recovery is a credible set that is sufficiently small to be informative, yet not so small that it does not cover the true parameter. The goal is to have a posterior distribution that contracts to its center at the same rate at which the estimator approaches the true parameter value.  More formally, the prior gives rise to posterior contraction if the posterior mass of the set $\{w: ||w-w_0||^2 \ge M p_n \log(n/p_n)\}$ converges to zero, where $w_0$ is the true parameter, $p_n=o(n)$, and $M$ is a constant. The landmark article   \cite{ghosal2000convergence} shows that the posterior concentration property at a particular rate implies the existence of a frequentist estimator that converges at the same rate. Consequently, if the posterior contraction rate is the same as the optimal frequentist estimator, the Bayes procedure also enjoys optimal properties.

For example, consider the Laplace prior, as used in a Bayesian approach to the Lasso, it is well-known that the Laplace distribution with rate parameter $\lambda$ can be represented as a scale mixture of normals where the mixing density is exponential with parameter $\lambda^2/2$ \cite{armagan}, however Theorem 7 in \cite{castillo2015bayesian} shows that if the true vector is zero, the posterior concentration rate shown in the full posterior does not shrink at the minimax rate. Theorem 2 gives condition under which the prior induced by the HALO penalty exhibits posterior contraction. The proof is given in the supplementary material.


\begin{theorem}
    Let $\ell_0(p_n) = \left\{w: \#(w_i \neq 0) \leq p_n\right\}$. Define the prior induced by \eqref{eq:adph_reg} as $\pi_{HALO}(w)$ and the corresponding posterior distribution $\pi_{HALO}(w | x)$.  Define the event $A(w)= \{w : |w - w_0\|^2 > M_np_n \log(n/p_n)\}$. Let $\tau_n=(p_n/n)^\alpha$ with $\alpha>1$ or $\tau_n=(p_n/n)\left(\log(n/p_n)\right)^\frac{1}{2}$.
    If $\tau_n \rightarrow 0$, $p_n \rightarrow \infty$ and $p_n = o(n)$ as $n \rightarrow \infty$ then 
    $$ \sup_{w_0 \in \ell_0(p_n)} \mathbb{E}_{w_0} \pi_{HALO}\left(A(w) | X\right)\rightarrow 0 $$ 
    \noindent for every $M_n \rightarrow 0$.
\end{theorem}

Recently \cite{wei2020contraction} has shown, in the setting of classification, that for logistic regression provided that a prior has a sufficient concentration near zero and has sufficiently thick tails, posterior concentrates near the true vector of coefficients at the described rate and with high posterior probability, only selects sparse vectors like a spike-and-slab point mass prior. \cite{polson2018posterior} considers spike-and-slab deep learning with ReLU activation as a fully Bayes deep learning architecture that can adapt to unknown smoothness. It also gives rise posteriors that concentrate around smooth functions at the near-minimax rate.

\section{Numerical Results}
\label{sec:res}

We present results to motivate and justify the use of HALO as a penalty for learning sparse deep neural networks. To do so, we perform numerical experiments on image recognition datasets where we apply sparsity to the convolutional (feature extraction) layers of the network, which aim to answer the questions

\begin{enumerate}
    \item Does learning to shrink model parameters improve model performance under highly sparse scenarios?
    
    \emph{Section \ref{sec:ic}-\ref{sec:od}: Yes. In nearly all of our experiments on image recognition tasks, the HALO penalty leads to models with higher accuracy than competing methods even for pre-trained models.}
    
    \item Does the HALO penalty also prevent overfitting in neural networks? 
    
    \emph{Section \ref{sec:overfit}: Yes. On datasets with label noise, HALO learns to ignore irrelevant samples reducing the generalization gap by over $40\%$ and improving performance by over $10\%$ over standard training with weight decay.}
    
    \item Does the HALO penalty induce a particular type of sparsity?
    
    \emph{Section \ref{sec:spars_type}: Yes. The HALO penalty is a monotonic penalty which learns both layer-wise sparsity and low-dimensional feature representations. }

\end{enumerate}

\begin{table*}[ht]
    \centering
    \resizebox{2\columnwidth}{!}{%
    \begin{tabular}{| c | c | c | c | c | c | c |}
            \hline
            & \multicolumn{3}{c|}{LeNet-300-100} & \multicolumn{3}{c|}{LeNet-5-Caffe}\\\hline
             Experiment &  Accuracy & Sparsity & Sparsity at Baseline  & Accuracy & Sparsity & Sparsity at Baseline \\\hline 
            Baseline & 98.57 $(\pm 0.04)$ & 0.6542 & 0.6542 & 99.24 $(\pm 0.08)$ & 0.5907 & 0.5907\\\hline
            Random Init Magnitude Pruning  & 98.23 $(\pm 0.03)$ & 0.95 & 0.9  & 98.91 $(\pm 0.15)$ & 0.95 & 0.8\\
            Lottery Ticket  & \textbf{98.44} $(\pm 0.13)$ & 0.95 & \textbf{0.95} & \underline{99.00} $(\pm 0.03)$ & 0.95 & 0.9 \\\hline
            $L_1$ & 98.29 $(\pm 0.04)$ & 0.95 & 0.9 & 98.96 $(\pm 0.17)$  & 0.95 & 0.9\\
            SWS   &  98.17 $(\pm 0.11)$     & 0.95 & 0.9  &  98.96 $(\pm 0.15)$   & 0.95 & 0.9   \\
            HALO & \underline{98.40} $(\pm 0.10)$ & 0.95 & 0.9 & \textbf{99.12} $(\pm 0.21)$ & 0.95 & \textbf{0.95}\\\hline
    \end{tabular}
    }
    \caption{Accuracy of sparsity-inducing regularization regularization and one-shot magnitude pruning based methods for LeNet-300-100 and LeNet-5-Caffe on MNIST. To obtain the reported sparsity results for the baseline, we threshold values for one of the runs at $0.01$.}
    \label{tab:lenet}
\end{table*}

\subsection{Experimental Setup}

In our experiments we evaluate the full model trained with weight decay (baseline), several pruning techniques: random initialization pruning \cite{liu2018rethinking}, lottery ticket hypothesis \cite{frankle2018lottery} and  GraSP \cite{wang2020picking}, a masked training approach DST \cite{liu2020dynamic}, and sparsity inducing penalties: Lasso ($\Omega_1$), MCP \cite{zhang2010nearly}, SWS \cite{strawderman2013hierarchical}, and the MAP estimate for HALO (\ref{eq:adph_reg}) on image recognition tasks for maintaining accuracy while inducing sparsity, and at high sparsity levels\footnote{95\% sparsity is used for for comparison, particularly \cite{liu2018rethinking}, and yields competitive performance with the baseline model.}

We define sparsity to be the proportion of zero weights in convolutional layers. For all classification results, unless otherwise stated the results represent an average of five runs and the error bars represent one standard deviation. Reported sparsity values are estimated from a single run of the model. Training details, hyperparameter\footnote{Although the two hyperparameters $\psi$ and $\xi$ can be absorbed into $\lambda$ and are not true hyperparameters, they are included as initializations to calibrate $\Omega(\cdot)$ to the same order of magnitude as $\mathcal{L}(\cdot)$ at the start of training.}  sensitivity and selection guidelines (for $\psi$ and $\xi$) are discussed in the supplementary materials.  We use standard benchmark networks and datasets for evaluating pruning methods from \cite{frankle2018lottery, liu2018rethinking}, which are expanded upon in the supplementary material. 

\subsection{Image Classification}
\label{sec:ic}

We present accuracy and sparsity ratios for sparse deep neural networks for image classification. Each model contains several fully-connected or convolutional layers for extracting features which we prune, and a final classification layer that outputs the probabilities for each class. 

\subsubsection{Feedforward Networks on MNIST}
\label{sec:ffm}

In the first experiment, we evaluate on the MNIST dataset for digit classification with a fully-connected LeNet-300-100 network, and the convolutional LeNet-5 network by pruning all layers of the networks. Results benchmarking pruning techniques and sparsity-inducing penalties are summarized in Table~\ref{tab:lenet}\footnote{The reported sparsity is not the highest sparsity ratio attainable by the baseline models, rather a reference that some weights are small enough to threshold. At 0.95 sparsity the full model predicts randomly.}.  All methods perform similarly with the baseline model, and HALO and the LTH perform slightly better on both networks. Most methods  retain the baseline accuracy at over 0.8 sparsity.

\subsubsection{Convolutional Networks on CIFAR-10 and CIFAR-100}
\label{sec:cnn_cifar}

\begin{table*}[ht]
    \centering
    \resizebox{2\columnwidth}{!}{%
    \begin{tabular}{| c | c | c | c | c | c | c |}
            \hline
             & \multicolumn{3}{c|}{VGG-Like CIFAR-10}  & \multicolumn{3}{c|}{ResNet-50 CIFAR-10}  \\\hline
             Experiment &  Accuracy & Sparsity & Sparsity at Baseline  & Accuracy & Sparsity & Sparsity at Baseline \\\hline 
            Baseline & 93.76 $(\pm 0.20)$ & 0.2176 & 0.2176 & 93.48 ($\pm 0.12)$  & 0.1626 & 0.1626\\\hline
            GraSP & $92.52 (\pm 0.10)$ & 0.95 & 0.9 & $88.95 (\pm 0.16)$ & 0.95 & 0.25 \\
            Random Init Magnitude Pruning  & $93.05 (\pm 0.21)$ & 0.95 & 0.9 & $88.59 (\pm 0.09)$ & 0.95 & 0.25\\
            Lottery Ticket  & $93.18 (\pm 0.12)$ & 0.95 & 0.9 & 88.75 $(\pm 0.18)$ & 0.95 & 0.4\\\hline
            DST & $93.27 (\pm 0.13)$ & 0.95 & \underline{0.92} & $\underline{89.64} (\pm 0.23)$ & 0.95 & 0.17 \\\hline
            $L_1$ & \underline{93.51} $(\pm 0.11)$ & 0.95 & 0.9 & 89.30 $(\pm 0.46)$ & 0.95 & \textbf{0.6}\\
            MCP & $92.28 (\pm 0.07)$ & 0.95& 0.85 &  89.54 $(\pm 0.24)$) & 0.95 & 0.4\\\hline
            SWS   & $93.50 (\pm  0.15)$ & 0.95 & 0.9   & 88.67 $(\pm 0.29)$  & 0.95 & 0.5   \\
            HALO & \textbf{93.61} $(\pm 0.16)$ & 0.95 & \textbf{0.95} & \textbf{90.71} $(\pm 0.17)$ & 0.95 & \underline{0.55}\\\hline
            \hline
            & \multicolumn{3}{c|}{VGG-Like C-100}  & \multicolumn{3}{c|}{ResNet-50 C100} \\\hline 
            Experiment &  Accuracy & Sparsity & Sparsity at Baseline  & Accuracy & Sparsity & Sparsity at Baseline \\\hline 
             Baseline & 73.41 $(\pm 0.23)$ & 0.4836 & 0.4836 & 70.75 $(\pm 0.28)$ & 0.1018 & 0.1018\\\hline
            GraSP & $69.5 (\pm 0.38)$ & 0.95& 0.65 &  $57.76 (\pm 0.39)$  & 0.95 & 0.25\\
            Random Init Magnitude Pruning  & 70.57 $(\pm 0.37)$ & 0.95 & 0.65 & 60 .82 $(\pm 0.63)$ & 0.95 & 0.3\\
            Lottery Ticket  & 70.55 $(\pm 0.23)$ &  0.95 & \underline{0.8} & 61.09 $(\pm 0.35)$ &  0.95 & 0.25\\\hline
            DST & $ \underline{70.93} (\pm 0.29)$ & 0.95 & 0.77 & $ 62.49 (\pm 0.40)$  & 0.90 & 0.25 \\\hline        
            $L_1$ & 70.67 $(\pm 0.24)$ & 0.95 & 0.7  & 60.97 $(\pm 0.49)$ & 0.95 & 0.45\\            
            MCP & 70.81 $(\pm 0.37)$ & 0.95& 0.7 & \underline{64.51} $(\pm 0.43)$ & 0.95 & 0.35 \\\hline
            SWS   &   68.05 $(\pm 0.28)$    & 0.95 & \underline{0.8} & 56.22 $(\pm 1.09)$    & 0.95 & \underline{0.35}   \\
            HALO & \textbf{72.48} $(\pm 0.24)$ & 0.95 & \textbf{0.85} & \textbf{65.01} $(\pm 0.34)$ & 0.95 & \textbf{0.7}\\\hline
    \end{tabular}
    }
    \caption{Accuracy of sparsity-inducing regularization and one-shot magnitude pruning based methods for the VGG-like and ResNet-50 on CIFAR-10 and CIFAR-100. We threshold values at $0.01$ for ResNet-50 and $0.001$ for VGG-like. Results from baseline and pruning methods at $95\%$ sparisty are taken from \cite{liu2018rethinking}. For sparsity at baseline results, we compute the highest sparsity ratio that achieves performance within $0.15$ of the baseline.}
    \label{tab:c10_c100}
\end{table*}

\begin{figure}[ht]
    \centering
    \begin{subfigure}[b]{0.20\textwidth}
        \includegraphics[width=\textwidth]{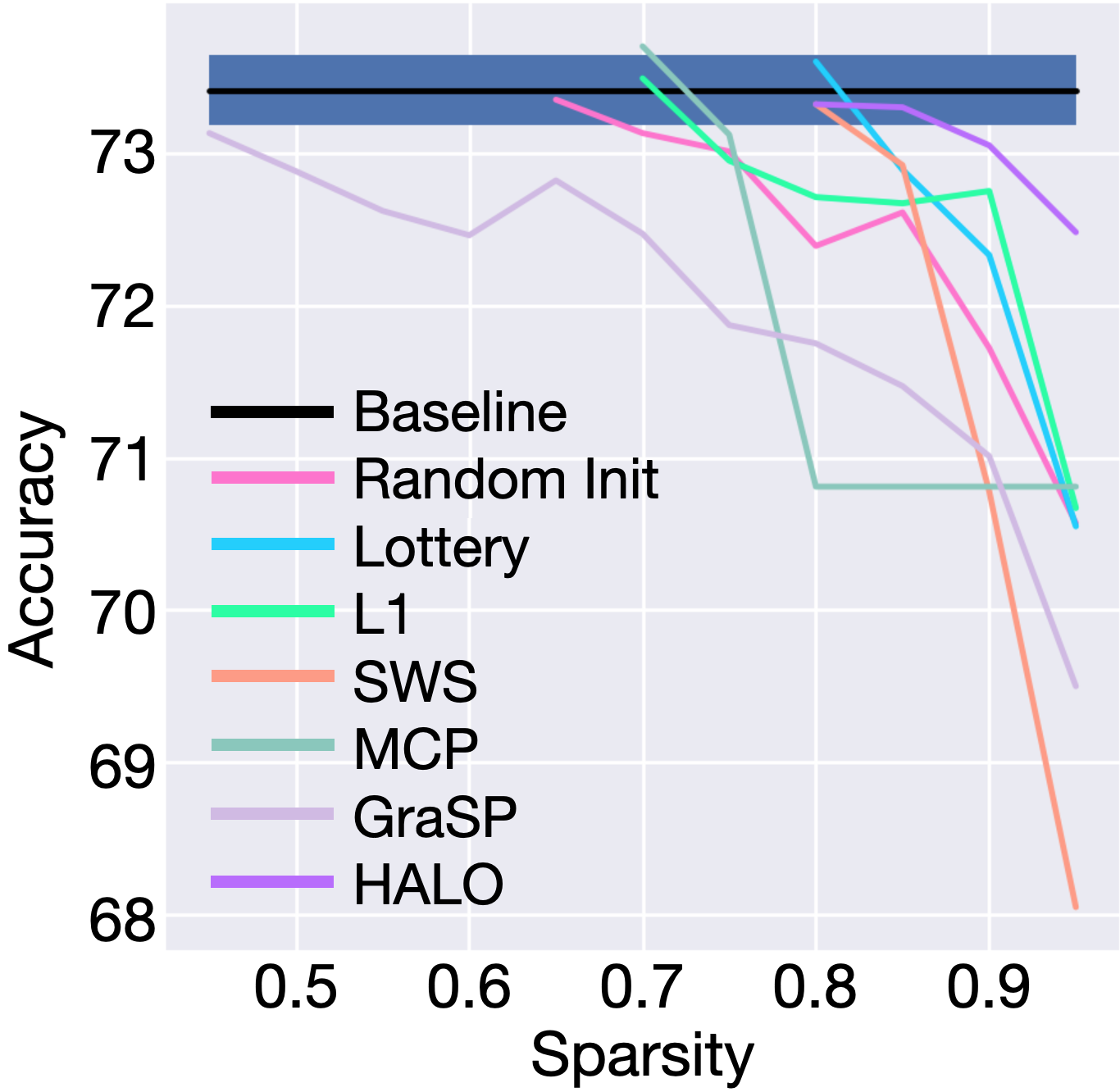}
        \caption{VGG on C100}
    \end{subfigure}
    \begin{subfigure}[b]{0.22\textwidth}
        \includegraphics[width=\textwidth]{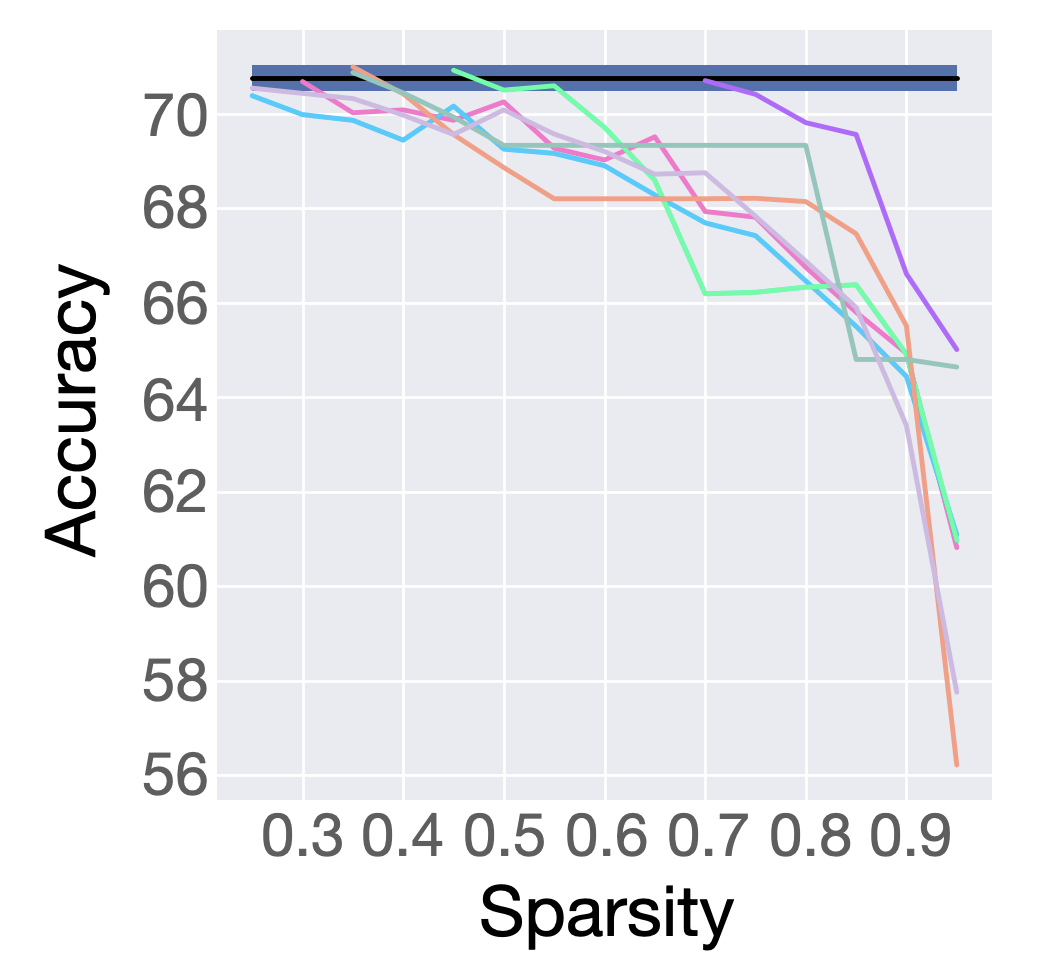}
        \caption{Res50 on C100}
    \end{subfigure}
    \caption{Visualization of test accuracy for various methods at different sparsity ratios. The solid black line represents the mean baseline accuracy over 5 runs and the blue region indicated one standard deviation.}
    \label{fig:other_sparsity}
\end{figure}

We additionally evaluate the VGG-like network and ResNet-50 architecture of \cite{liu2018rethinking} on the CIFAR-10/100 classification task. Results are summarized in Table~\ref{tab:c10_c100}. For VGG, regularization approaches outperform pruning methods and HALO is able to retain accuracy at high sparsity. On CIFAR-100 in particular, HALO drops performance by only $1\%$ compared with other methods that drop accuracy by $2-3\%$. On ResNet, all methods drop accuracy, however HALO performs the best at 0.95 sparsity, and achieves similar accuracy to the full model at comparable or higher sparsity ratios. Results at varying levels of sparsity are given in Figure~\ref{fig:other_sparsity} for CIFAR-100.  It is important to note, training with HALO always yields a model that attains competitive accuracy with 0.55 or greater sparsity.  Overall, methods with sparsity inducing penalties tend to have comparable or better performance than other types of methods, and HALO allows more flexible regularization compared to other sparsity inducing penalties.
The results illustrate that the flexibility of HALO allowing the model to learn to simultaneously train and sparsify leads to better results than learning independently or with less flexibility.

\subsection{Object Detection}
\label{sec:od}

\begin{figure}[h]
    \centering
    \includegraphics[height=1.5in,width=1.5in]{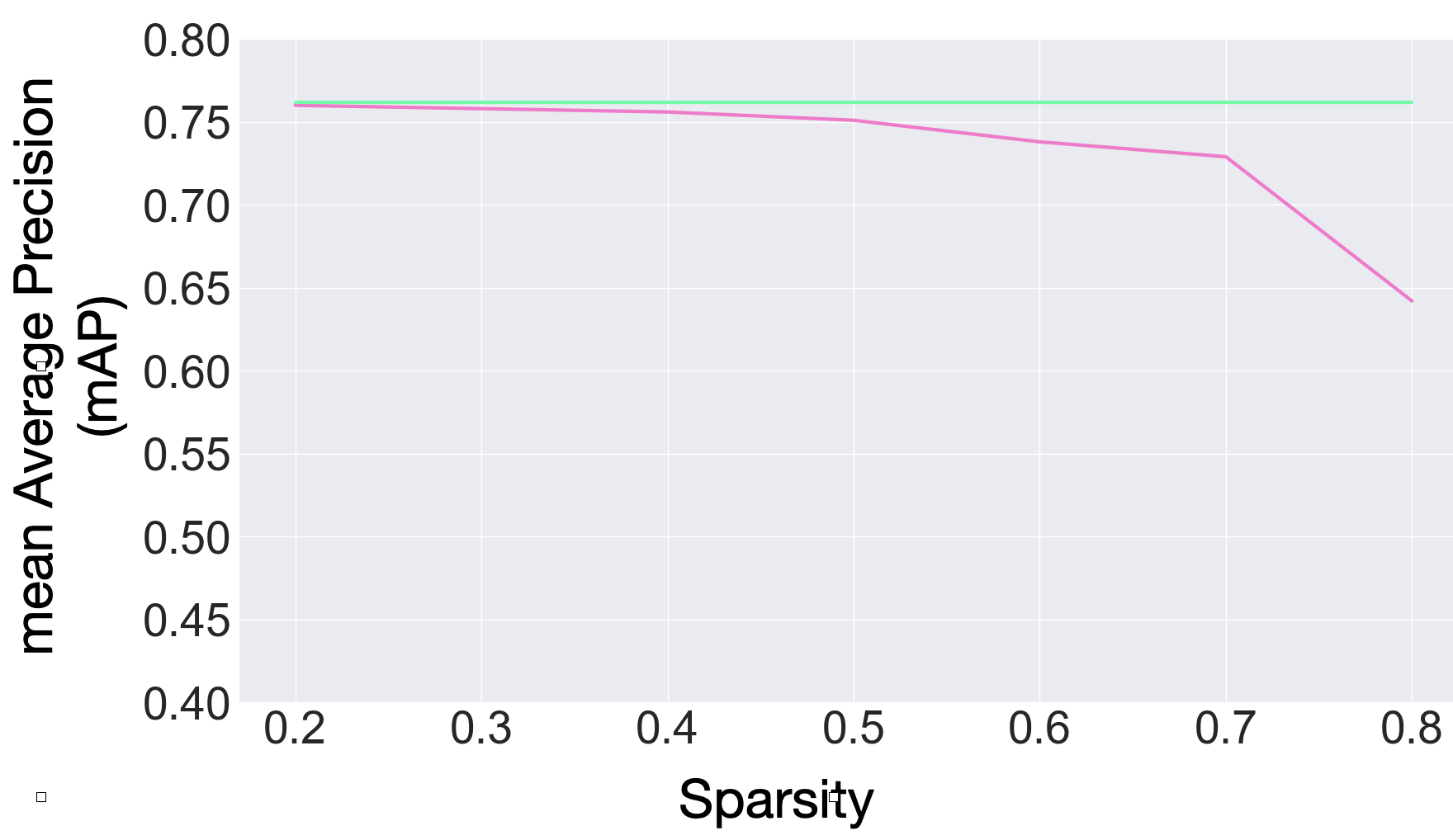}
    \caption{Mean Average Precision of SSD-300 models pruned by HALO (pink) vs. the baseline model (green).}
    \label{fig:od}
\end{figure}

We investigate HALO performance for transfer learning to other image recognition tasks.  We demonstrate this using the SDS-300 network \cite{liu2016ssd} on the PASCAL VOC  object detection task where the goal is to classify objects in an image and estimate bounding box coordinates. In this setting,  a VGG network is trained to classify ImageNet,  the classification layers are removed, additional convolutional layers for predicting object bounding boxes are defined, and the network is fine-tuned on the Pascal VOC dataset.  When fine-tuning the network with the HALO penalty, the network can be pruned significantly while maintaining similar accuracy as shown in Figure~\ref{fig:od}. We find that at 50\% sparsity, mean average precision (mAP) drops by  0.013 mAP averaged over 5 runs, and at 70\% sparsity, performance only drops by around 0.03 mAP indicating that feature extraction capability from pre-training has been preserved.

\subsection{Regularization for Limiting Overfitting}
\label{sec:overfit}

\begin{figure}
    \centering
    \begin{subfigure}[b]{0.16\textwidth}
        \includegraphics[width=\textwidth]{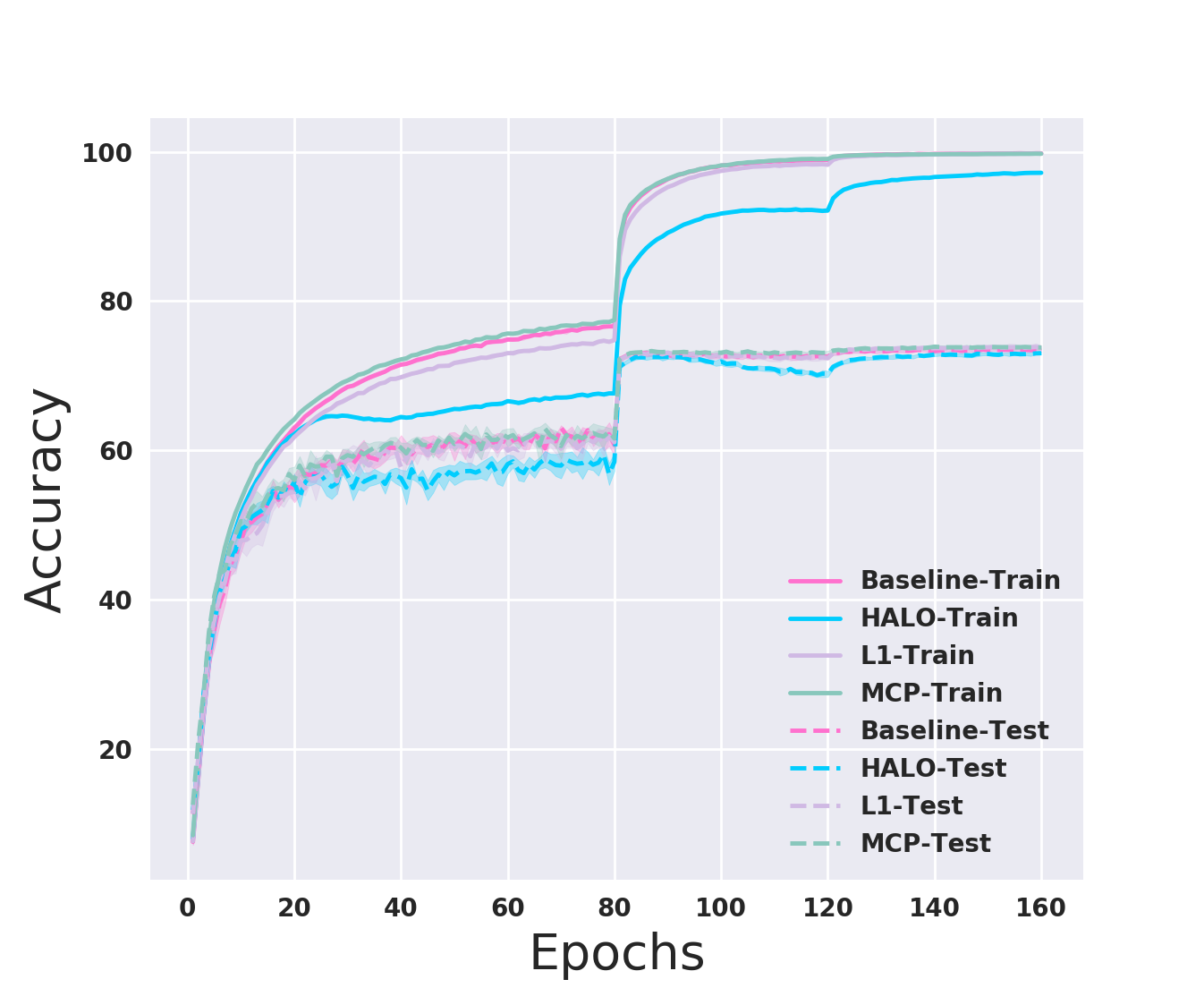}
        \caption{$\rho=0$}
    \end{subfigure}
    \begin{subfigure}[b]{0.16\textwidth}
        \includegraphics[width=\textwidth]{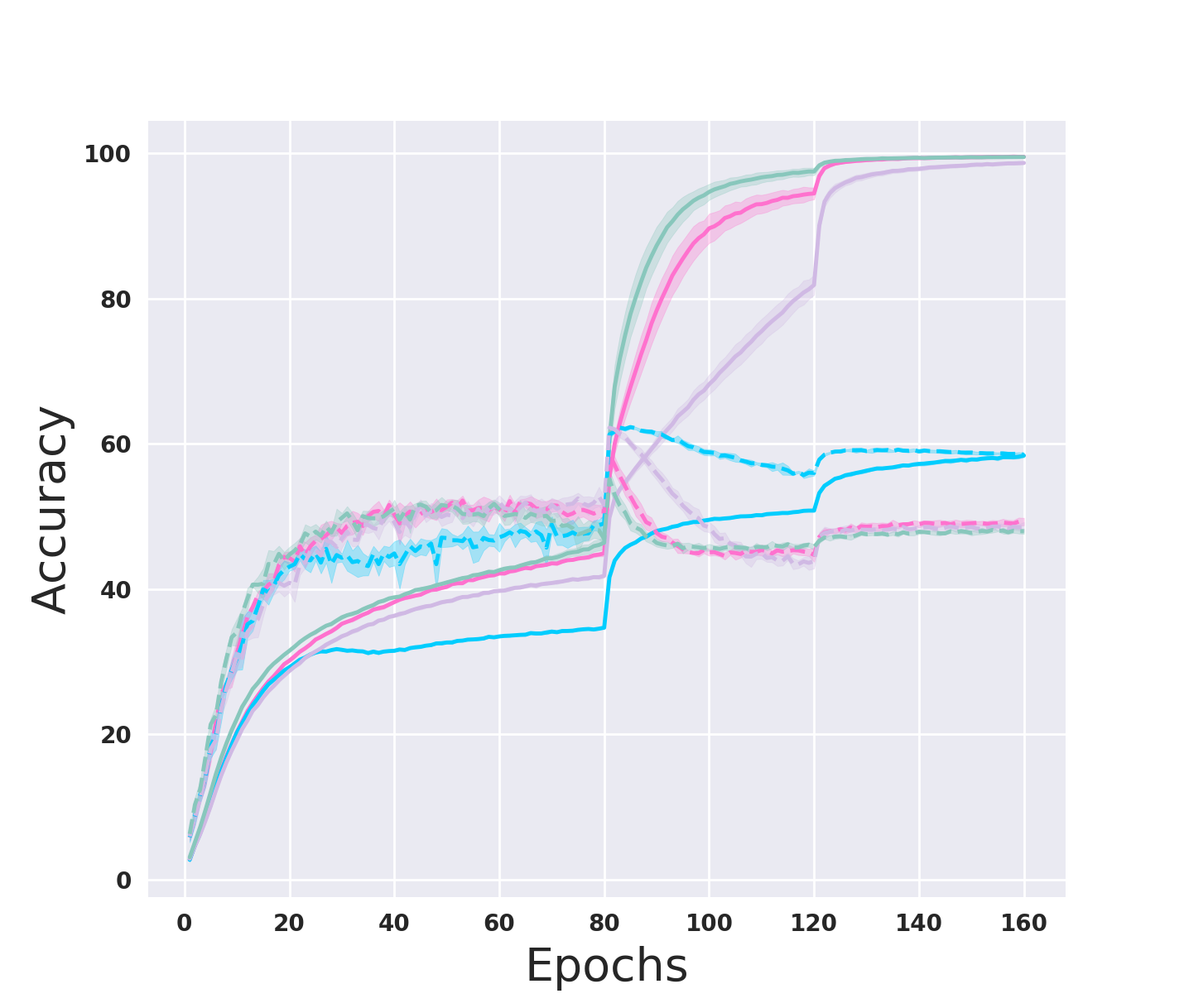}
        \caption{$\rho=0.4$ }
    \end{subfigure}
    \begin{subfigure}[b]{0.15\textwidth}
        \includegraphics[width=\textwidth]{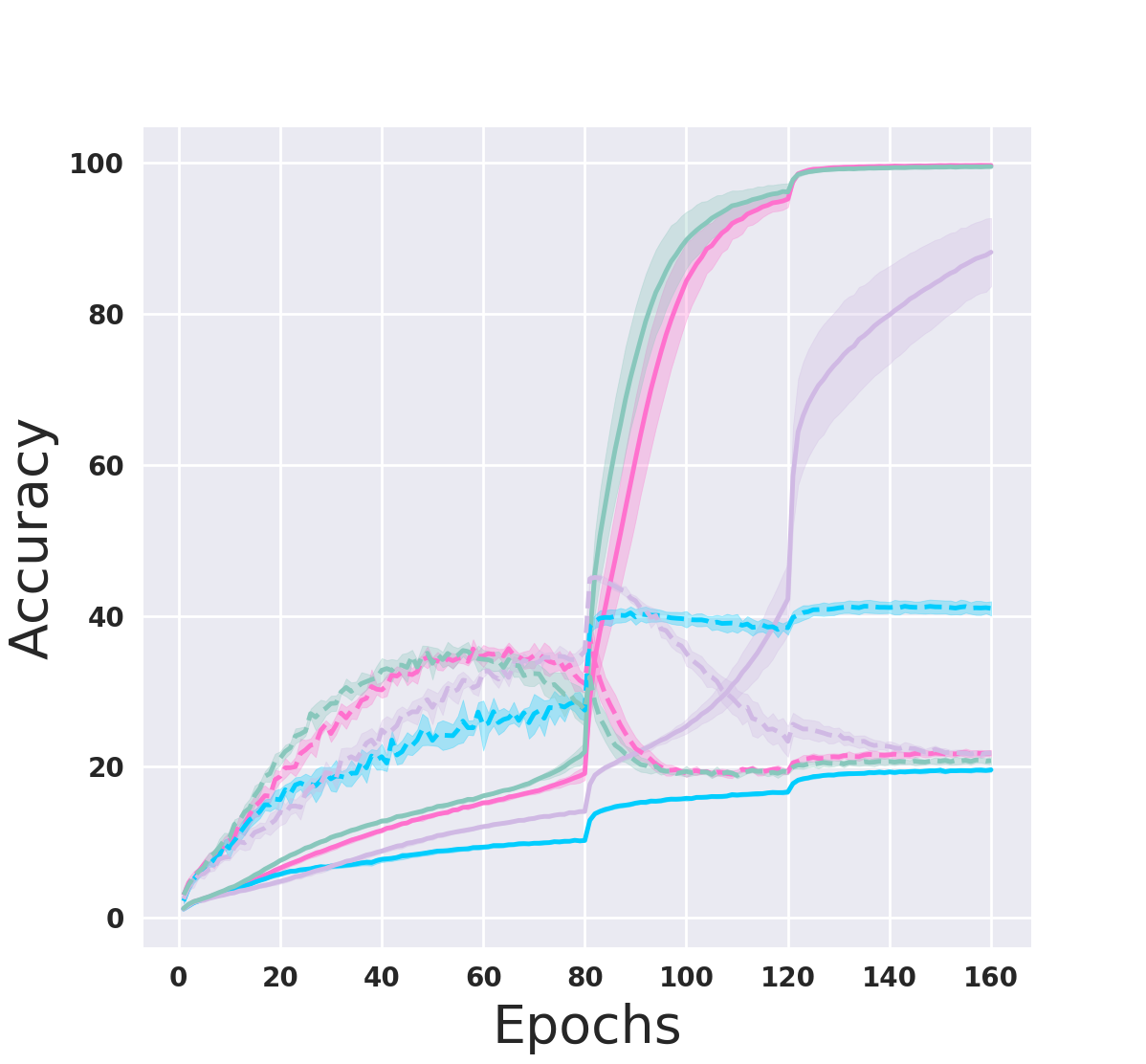}
        \caption{$\rho=0.7$}
    \end{subfigure}
    \caption{Training and test errors for each epoch of training with the sparse regularizers and the baseline model with weight decay. The mean accuracy over 5 runs and one standard deviation are shown at varying amounts of label noise $\rho$.}
    \label{fig:gen}
\end{figure}

In addition to inducing sparsity and pruning neural network architectures, regularization is a tool for limiting overfitting to noisy data \cite{ahmad2019can, saxena2019data, zhang2016understanding}. A common setting where overfitting can occur is in the presence of label noise where the label of each image in the training set is independently changed to another class label with probability $\rho$. We implement training of the VGG-like architecture with varying label noise on CIFAR-100. Figure~\ref{fig:gen} shows the training accuracy for the baseline model approaches $100\%$ while the test error does not when $\rho>0$.  For the VGG-like model trained with the HALO penalty (using suitable hyperparameters to match test accuracy at $\rho=0$), we find that at $\rho=0.0$, both train and test curves follow nearly identical patterns, whereas at $\rho=0.4, 0.7$ the training accuracy does not increase to $100\%$, but reaches similar accuracy proportional to the clean images. At $\rho=0.4$, the test error is roughly the same as the training error indicating a small generalization gap unlike with standard training, while for $\rho=0.7$ the test accuracy is higher indicating underfitting due to the lack of data. These results indicate HALO can diagnose mis-labeled data in the training set.

\subsection{Learning Different Types of Sparsity}
\label{sec:spars_type}

\begin{figure}[h!]
    \centering
    \begin{subfigure}[b]{0.145\textwidth}
        \includegraphics[width=\textwidth]{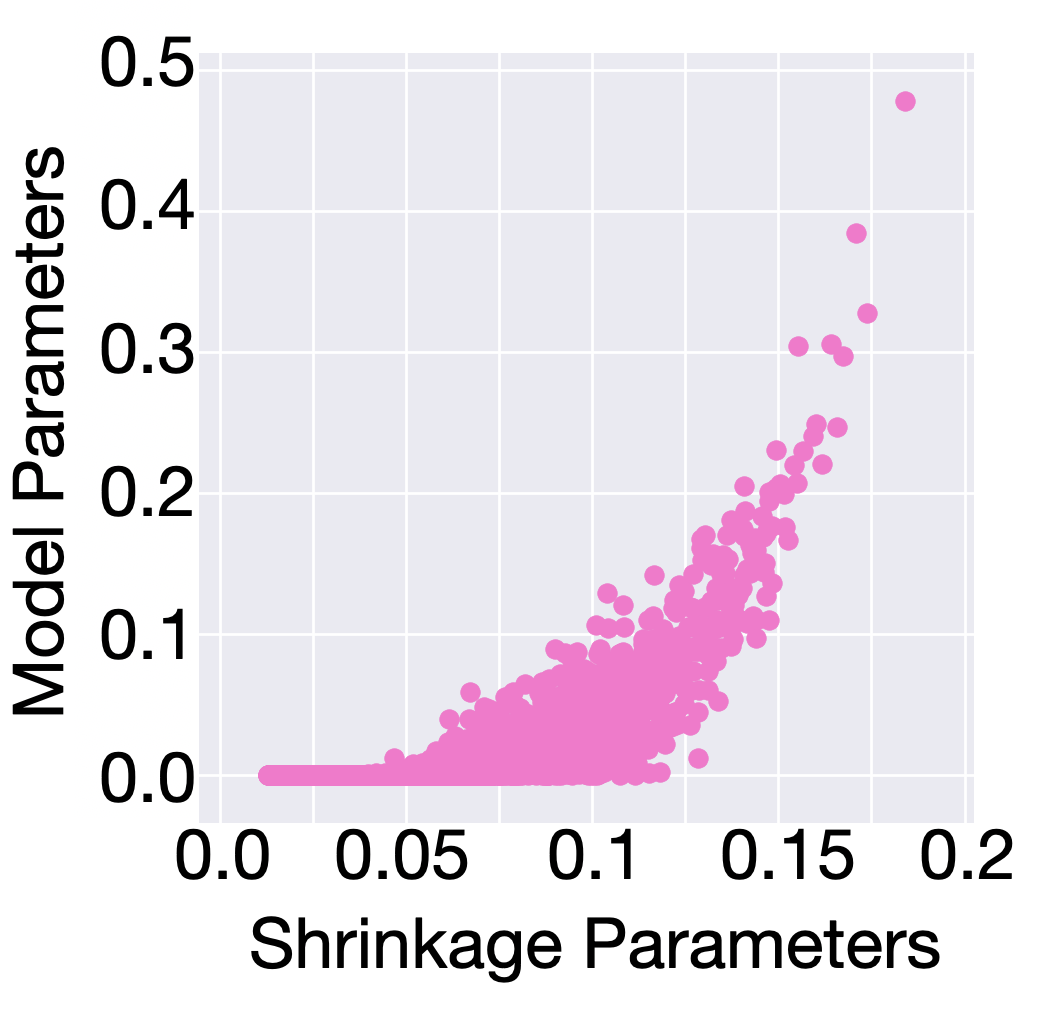}
        \caption{Monotonic}
        \label{fig:sp}
    \end{subfigure}
    \begin{subfigure}[b]{0.145\textwidth}
        \includegraphics[width=\textwidth]{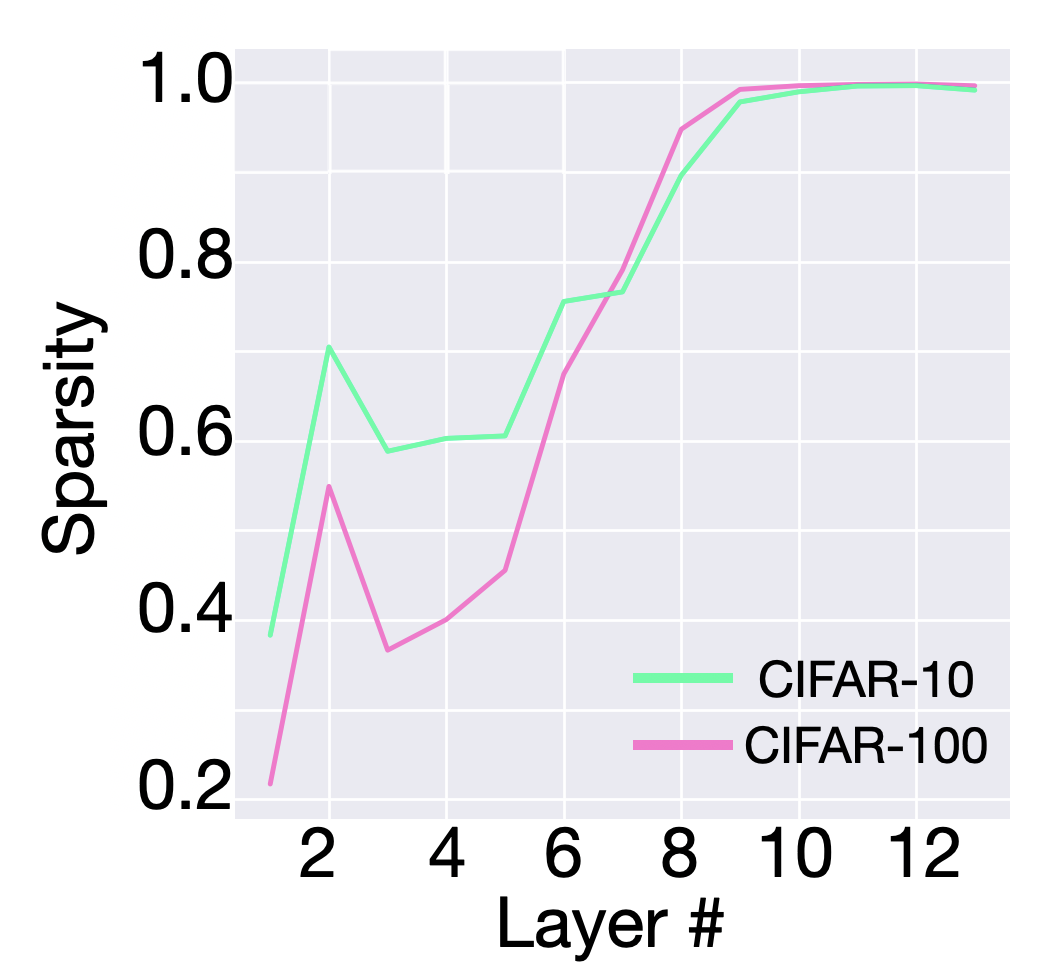}
        \caption{Structured }
         \label{fig:stp}
    \end{subfigure}
    \begin{subfigure}[b]{0.16\textwidth}
        \includegraphics[width=\textwidth]{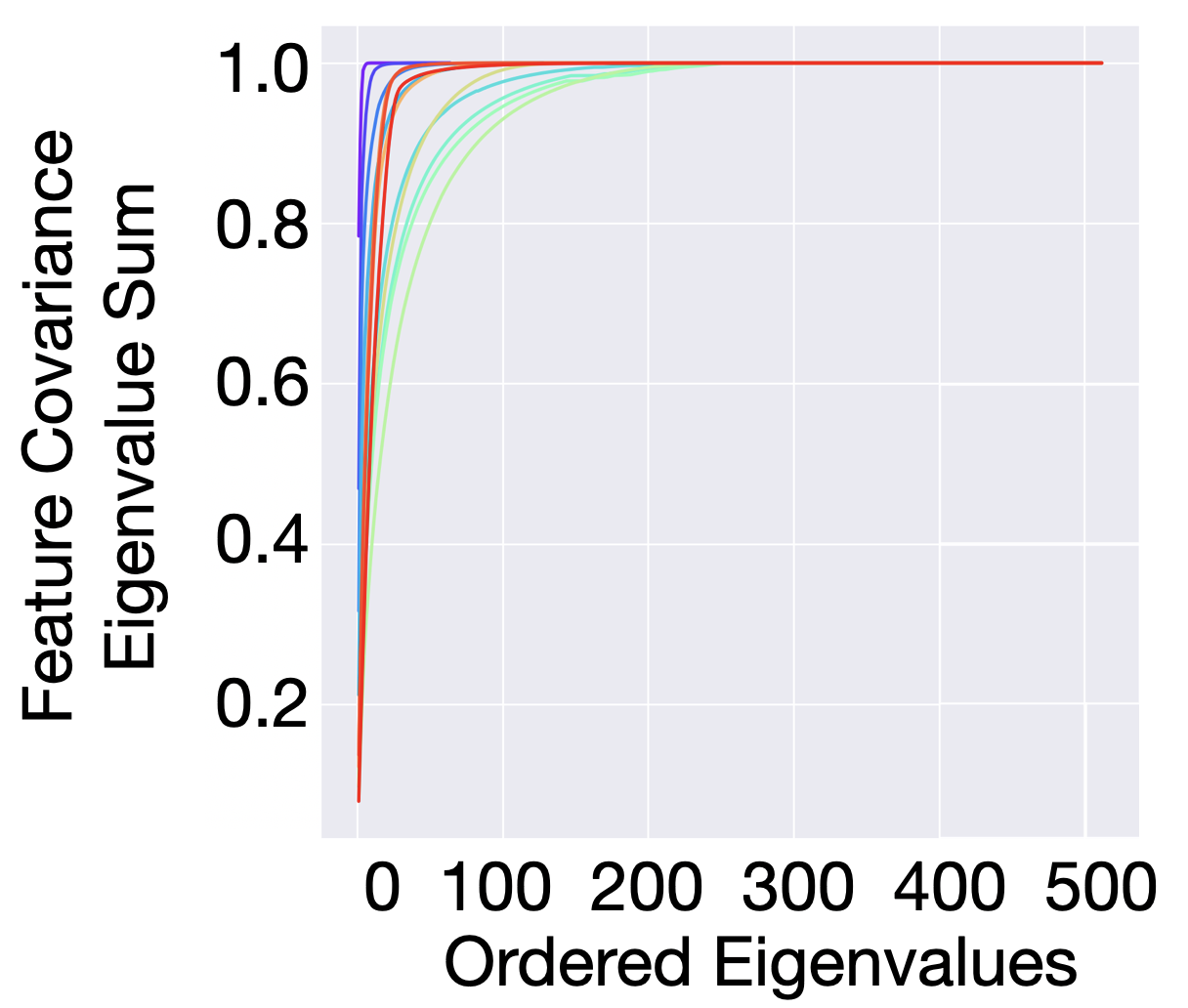}
        \caption{Low-rank}
    \end{subfigure}
    \caption{Different types of sparsity learned by HALO. (a) Monotonic trend of a random sample of 10,000 regularization coefficients and weight parameters. (b) Sparsity by layer for CIFAR-10/0 indicating pruning of entire layers. (c) Normalized cumulative sum of eigenvalues of the feature covariance matrix indicating a small number of important features \cite{Cuadros_2020_WACV}. Color range from violet (low) to red  (high) indicates layer number.}
    \label{fig:pen}
\end{figure}

We have shown that our approach can prune different network architectures in order to achieve small models with little drop in performance.  We present results to highlight thorough exploration on the VGG-like architecture that the HALO penalty performs monotonic penalization, learns structured sparsity, and learns low-rank feature representations leading to faster networks with smaller memory footprints. Results are summarized in Figure~\ref{fig:pen} and expanded upon in \ref{sec:spars_type}.

\section{Conclusion}
\label{sec:con}

Over-parameterization is a challenging problem preventing the use of deep neural networks for learning representations in devices with limited computational budgets. In this work, we present HALO, a novel penalty function which when used to train neural networks produces subnetworks which achieve state-of-the-art performance compared with magnitude criteria pruning techniques, without re-training the subnetwork. Our approach has several benefits. It is simple to implement and does not require storing model weights or re-training unlike other network pruning methods. While the method is limited by additional  additional training parameters from the regularizer, this is common to many other approaches \cite{azarian2020learned, kusupati2020soft, liu2020dynamic} and no additional parameters need to be stored after training.  Further, our approach can be combined with other loss functions, is not limited to classification, and is model agnostic. We believe that this work is one step towards creating penalty functions that can be applied to sparsify any model without performance degradation. Further research in this area can lead to impressive performance gains in settings with limited computational resources, and towards a better understanding of the rich feature representations learned by neural networks.

\section*{Acknowledgments}
We thank Ben Baer for his helpful advice and discussions regarding the LLA algorithm and one-step estimators in Section~\ref{sec:method}, as well as general pointers for optimization with sparsity-inducing penalties and shrinkage priors. Wells' research supported by NIH grant R01GM135926.

\bibliographystyle{plain}
\bibliography{bib}

\clearpage
\section{Appendix}

\subsection{Generalizations of HALO}

\subsubsection{Other choices of h(x)}
\label{sec:oh}

We introduced the HALO penalty and in our experiments we consider $h(x) = \frac{1}{x^2}$ to enforce $h(x) \rightarrow \infty$ as $x \rightarrow 0$. We found that this combined with the $L_1$ penalty on $\lambda_j$ encourages selective shrinkage of the weights where important weights remain unregularized. Using $h(x) = \frac{1}{x^k}$ when $k>0$ is favorable because $h(x) \rightarrow 0$ and $\infty$ as $x \rightarrow \infty$ and $0$ respecitvely. With this choice of $h(x)$ our penalty is a flexible variant of the adaptive penalties such as magnitude pruning or the relaxed Lasso because the regularization coefficients in these approaches (regularization coefficients are pre-specified as either zero or infinity) are the limit of those learned by HALO. However in some cases it may not be favorable to use a sharp penalty on the regularization coefficients since this may lead to over-sparsification of the weights.

\begin{figure}[ht]
    \centering
    \begin{subfigure}[b]{0.2\textwidth}
        \includegraphics[width=\textwidth]{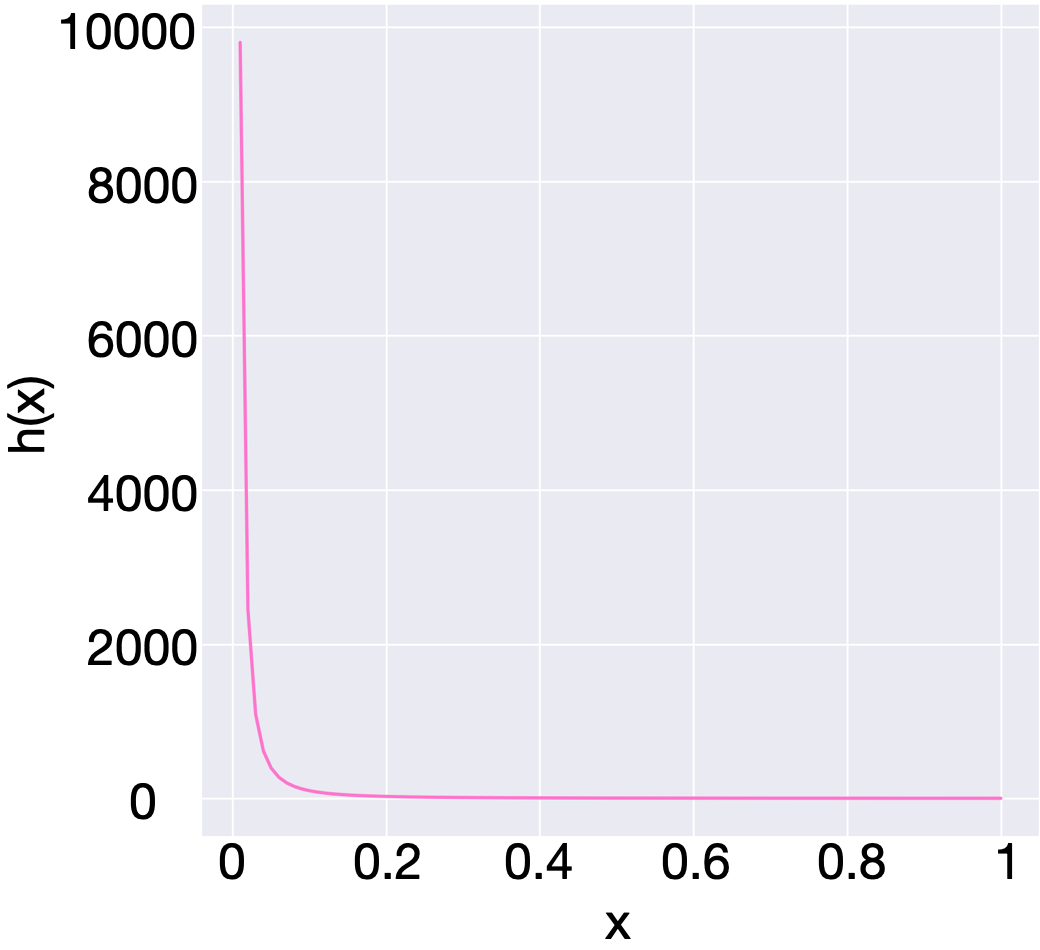}
        \caption{$\frac{1}{x^2}$}
    \end{subfigure}
    \begin{subfigure}[b]{0.2\textwidth}
        \includegraphics[width=0.95\textwidth]{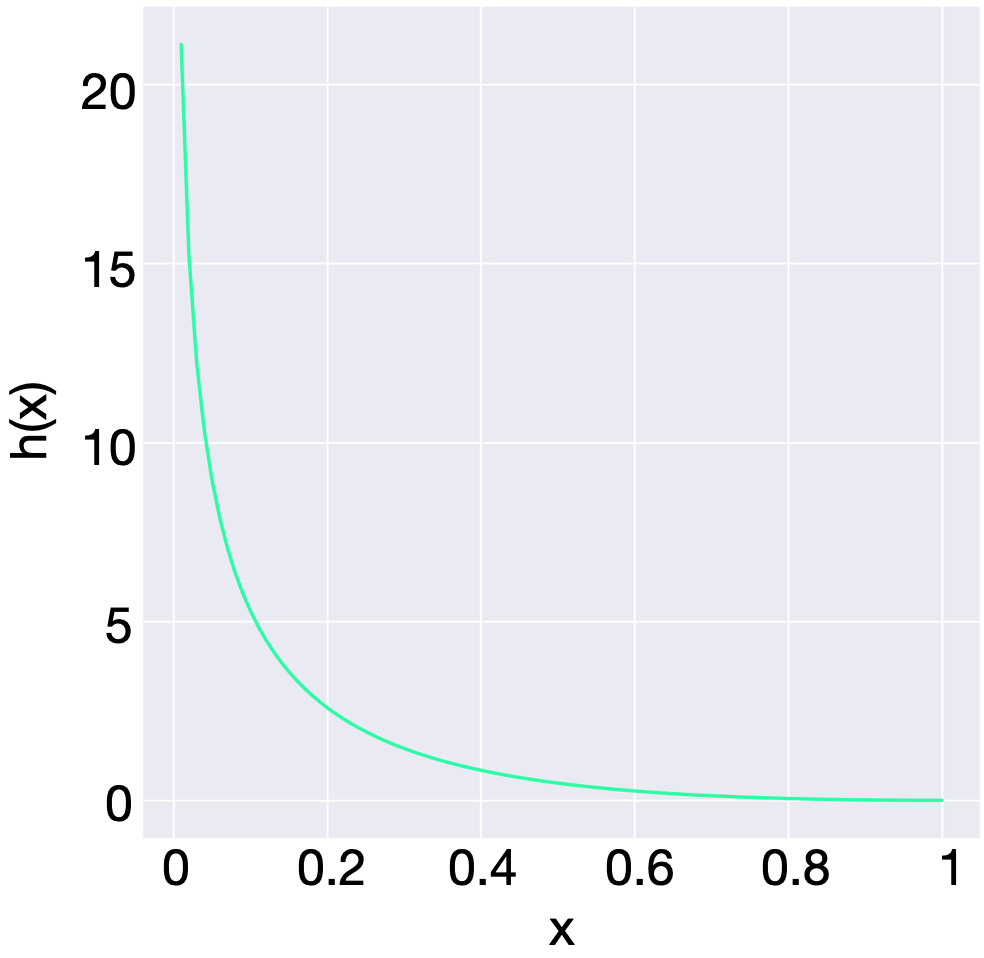}
        \caption{$\log(|x|)^2$}
    \end{subfigure}
    \caption{Comparing different choices of $h(x)$ for HALO.}
    \label{fig:ofunc}
\end{figure}

As an alternative, we investigate $h(x) = \log(|x|)^2$.  Figure~\ref{fig:ofunc} highlights the key difference: $\log(|x|)^2$ approaches infinity at a much slower rate over  $\frac{1}{x^2}$, thus inducing less shrinkage of small coefficients. Another difference is that for $x>1$, $\log(|x|)^2$ is an increasing function. This means that increasing $\lambda$ will also penalize weights, although this will rarely happen since the $L_1$ penalty acts to shrink the $\lambda$s as much as possible.  On experiments with VGG-16 on CIFAR-100 we obtain comparable accuracy of $72.36 \pm 0.22$ for VGG-like on CIFAR-100. This is similar to the performance with $h(x) = \frac{1}{x^2}$ ($72.48 \pm 0.24$).

Although we have not fully explored the range of possible functions for HALO, we believe that the choice of $h(x)$ will be data, model and application dependent. A choice of $h(x)$ such that $h(x) \rightarrow \infty$ as $x \rightarrow 0$ and $0$ as $x \rightarrow \infty$ will lead to a soft-thresholding variant of the common magnitude pruning approaches, while a choice of $h(x) \rightarrow \infty$ as $x \rightarrow 0$ and $\infty$ will impose additional restraints on weight magnitudes (favoring medium magnitude weights). Still there are many other functions we can explore in future work including piecewise functions that have different behaviors for large and small weights, or those which lead to other penalty behaviors such as sorted penalties \cite{bogdan2013statistical, feng2019sorted} which penalize larger weights more strongly.

\subsubsection{Structured HALO}
\label{sec:sh}

We demonstrate that HALO learns efficient sparse architectures in performing effective dimensionality reduction of the features and nearly sparsifying entire layers. However, HALO is not a structured penalty and it is not guaranteed to enforce sparsity at a group level.  While prior work \cite{malach2020proving} shows that subnetworks pruned at the neuron-level are unable to attain performance equivalent to trained networks, whereas subnetworks pruned at the weight-level can, in some cases structured sparsity may be more desirable. We propose a natural extension of HALO to structured penalization for deep neural networks. One possible version of a structured HALO (SHALO) penalty is based on the composite penalty framework proposed in \cite{breheny2009penalized}

\vspace{-3mm}
$$\Omega_C = \Omega_O\left(\sum_{j=1}^{p_g} \Omega_I(|w_{gj}|)\right),$$
\vspace{-3mm}

where $\Omega_O$ is some penalty applied to the sum of inner penalties $\Omega_I$ and $w_{gj}$ is the $j$th member of the $g$th group. This framework is general for group penalties and includes both the group bridge penalty and group Lasso \cite{huang2012selective}, and this approach can be applied with the HALO penalty for learning structured sparsity in deep neural networks. 


 We suggest two approaches for extending HALO

\begin{enumerate}
    \item Apply a Lasso penalty for the inner penalty and HALO for the outer penalty:
        $$ \Omega_{\text{SHALO}} = \xi\sum_{g=1}^G h(\lambda_g) \sum_{j=1}^{p_g} |w_{gj}| + \psi \sum_{g=1}^G|\lambda_g|,$$
        which learns regularization coefficients for controlling groups of weights only.
        
    \item Apply the HALO penalty for both the inner and outer penalty: 
        $$ \Omega_{\text{SHALO}} = \xi\sum_{g=1}^G h(\lambda_g) \sum_{j=1}^{p_g} \Omega_{HALO}(\mathbf{w_g}) + \psi \sum_{g=1}^G|\lambda_g|, $$
        where $\mathbf{w_g}$ is the vector of all weights in the $g$th group. This penalty will learn regularization coefficients group-wise and for individual weights.
\end{enumerate}

In future work, we hope to consider these penalties and other structured variants for learning more efficient sparse networks.

\subsection{Theorem 1: Convexity of HALO in the Linear Model}
\label{sec:conv}
We demonstrate that for the standard linear model, the loss function with the HALO penalty is convex. 

     \begin{lemma}
        \label{lemma:1}
        For any symmetric matrix $$M = \begin{bmatrix}
                 A & B\\
                 B & C
                 \end{bmatrix}$$
                 
        if $A$ is invertible, then $M \succ 0$ iff $A \succ 0$, and $C - B^TA^{-1}B \succ 0$.
    \end{lemma}
        
    We invoke Lemma 1 to show convexity in both $W$ and $\bm{\lambda} = [\lambda_1, \lambda_2, \dots, \lambda_p]$:

    \begin{proof}
        First, note that $\nabla^2 L(W, \bm{\lambda})$ is symmetric and can be written as a block matrix since
        
        \begin{align*}
            \nabla^2  L(W, \bm{\lambda}) &= \begin{bmatrix}
                 A & B\\
                 B & C
                 \end{bmatrix}\\
            &=  \begin{bmatrix}
                     \text{diag}(\frac{6\xi w_k}{\lambda_k^4}) &  \text{diag}(\pm\frac{2\xi}{\lambda_k^3})\\
                     \text{diag}(\pm\frac{2\xi}{\lambda_k^3}) & X^TX
                \end{bmatrix}
        \end{align*}
        
        Second, since $A$ is a diagonal matrix with positive values along the diagonal, $A$ is both invertible and $A \succ 0$. Consider the expression $$\Lambda = C - B^TA^{-1}B.$$

        Note that if $X$ is full rank, then the smallest eigenvalue of $X^TX$ is $\nu > 0$. Further, the eigenvalues of $B^TA^{-1}B$ are $$\rho_k = \frac{4\xi^2}{\lambda_k^6} \cdot \frac{6\xi |w_k|}{\lambda_k^4}.$$
        
        Then, By Weyl's theorem \cite{fan1949theorem},
        
        $$\nu - \rho_p \leq \mu_p \leq \nu - \rho_1$$
        
        where $\mu_p$ is the smallest eigenvalue of $\Lambda$, $\rho_p$ is the smallest eigenvalue of $B^TA^{-1}B$, and $\rho_1$ is the largest. Then, since $$24\nu \frac{\lambda_p^{10}}{|w_p|} \leq \xi^3 \leq 24\nu \frac{\lambda_1^{10}}{|w_1|}$$
        
        we have 
        
        $$0 \leq \nu - \rho_p \leq \mu_p \leq \nu - \rho_1$$
        
        and $\Lambda \succ 0$.
        
    \end{proof}
    
    \emph{Remark: Note that the conditions of the proof do not depend on $\psi$. While we believe $\psi$ cannot be entirely disregarded, based on the above theorem and empirically, HALO has only one key hyperparameter $\xi$ rather than two.}

\subsection{Theorem 2: Posterior Contraction of HALO}
\label{sec:posterior}

We demonstrate that the posterior contraction property holds by satisfying the conditions of \cite[Theorem 4]{ghosh2017}. The results in \cite{ghosh2017} use the notion of slowly-varying functions and their general theorem depends on the fact that the hierarchical prior can be represented in terms of such a function.
\cite{ghosh2017} assume the following two conditions hold for some slow-varying function $L(\cdot)$:
    
    \begin{enumerate}
        \item $\lim_{t \rightarrow \infty} L(t) \in (0, \infty)$ and
        \item there exists some $0 < M < \infty$ such that $\sup_{t \in (0, \infty)} L(t) < M$.
    \end{enumerate}
    
    Specifically, in \cite{ghosh2017} it is shown that a large class of mixture of normal priors, specifically normal-exponential-gamma priors, are in the family of ``Three Parameter Beta'' (TPB) priors \cite{armagan}. Membership in the TPB family of priors implies the normal-exponential-gamma prior can be represented by a function proportional to $w_i^{-\beta-1}L(w_i)$ for some $\beta > 0$ and a slowly varying function $L(\cdot)$ satisfying the assumptions of Theorem 2 \cite{ghosh2016}.  With the prior having the representation connected to the slowly vary function $L$, it is shown that the normal-exponential-gamma prior has the posterior contraction property. 

For the prior $\pi_{HALO}$, we can demonstrate posterior contraction because the tail properties of $\pi_{HALO}$ are the same as those of the normal-exponential-gamma prior. Therefore in terms of the tail behavior, it follows that $\pi_{HALO} \propto w_i^{-\beta-1}\tilde{L}(w_i)$ for some $\beta > 0$ and $\tilde{L}$ satisfying the conditions of Theorem 2.  Consequently, the posterior contraction of $\pi_{HALO}$ follows directly  from \cite[Theorem 4]{ghosh2017}.

\subsection{Training Configurations}
\label{sec:tc}

We use the standard train/test split for the MNIST digits dataset containing 60,000 training images and 10,000 test images, and CIFAR-10/0 datasets which contain 50,000 training images and 10,000 test images available from the torchvision dataloaders\footnote{\url{https://pytorch.org/docs/stable/torchvision/datasets.html}}. For the Pascal VOC dataset, we train using the trainset from Pascal 2007 and Pascal 2012, and test on the Pascal 2007 testset.

We train all models using SGD with a momentum of $\gamma = 0.9$ and weight decay. For MNIST, we use a batch size of $100$ and train with an initial learning rate of $0.1$ decaying by $0.1$ at every 25k batches for $250$ epochs, and use weight decay of $0.0005$. For CIFAR-10/100 we use a batch size of $64$, and train with an initial learning rate of $0.1$ decaying by $0.1$ at the 80th and 120th epochs for 160 epochs. We set the weight decay parameter to be $0.0001$. For CIFAR-10/100 experiments, we use standard data augmentations (random horizontal flip, translation by 4 pixels).  For Pascal VOC, we train with a batch size of 32 and weight decay 0.0005 for 120,000 steps at a learning rate of 0.001 decreasing the learning rate at 80,000 and 100,000 steps. Regularization coefficients are initialized at one for all $\lambda_j$ and have their own optimizer but follow the same decay rate. This also reduces our approach to Lasso for the first batch.

\subsection{Parameter Robustness}

\begin{figure}[ht]
    \centering
    \begin{subfigure}[b]{0.2\textwidth}
        \includegraphics[width=\textwidth]{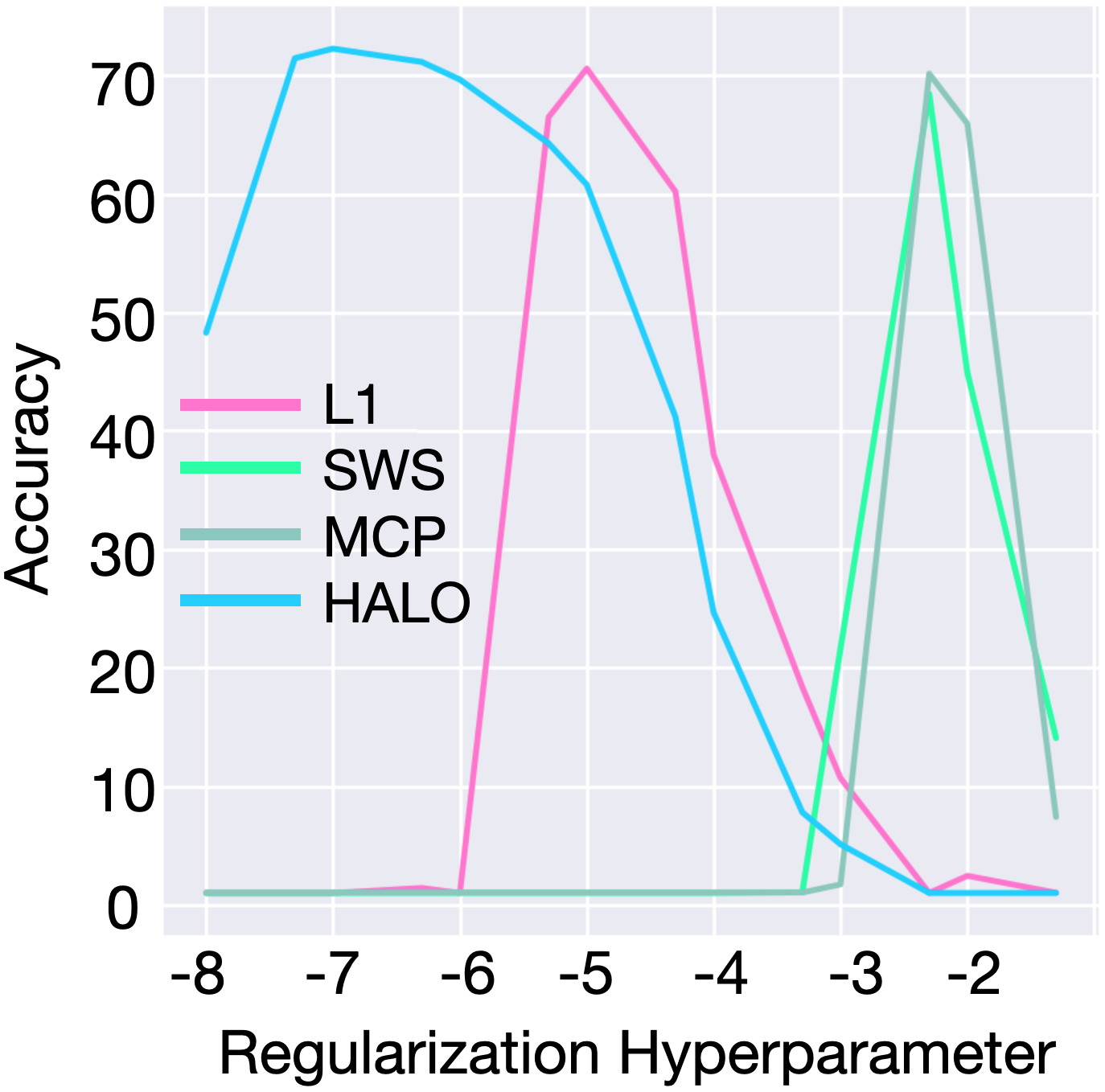}
        \caption{VGG-like}
    \end{subfigure}
    \begin{subfigure}[b]{0.2\textwidth}
        \includegraphics[width=\textwidth]{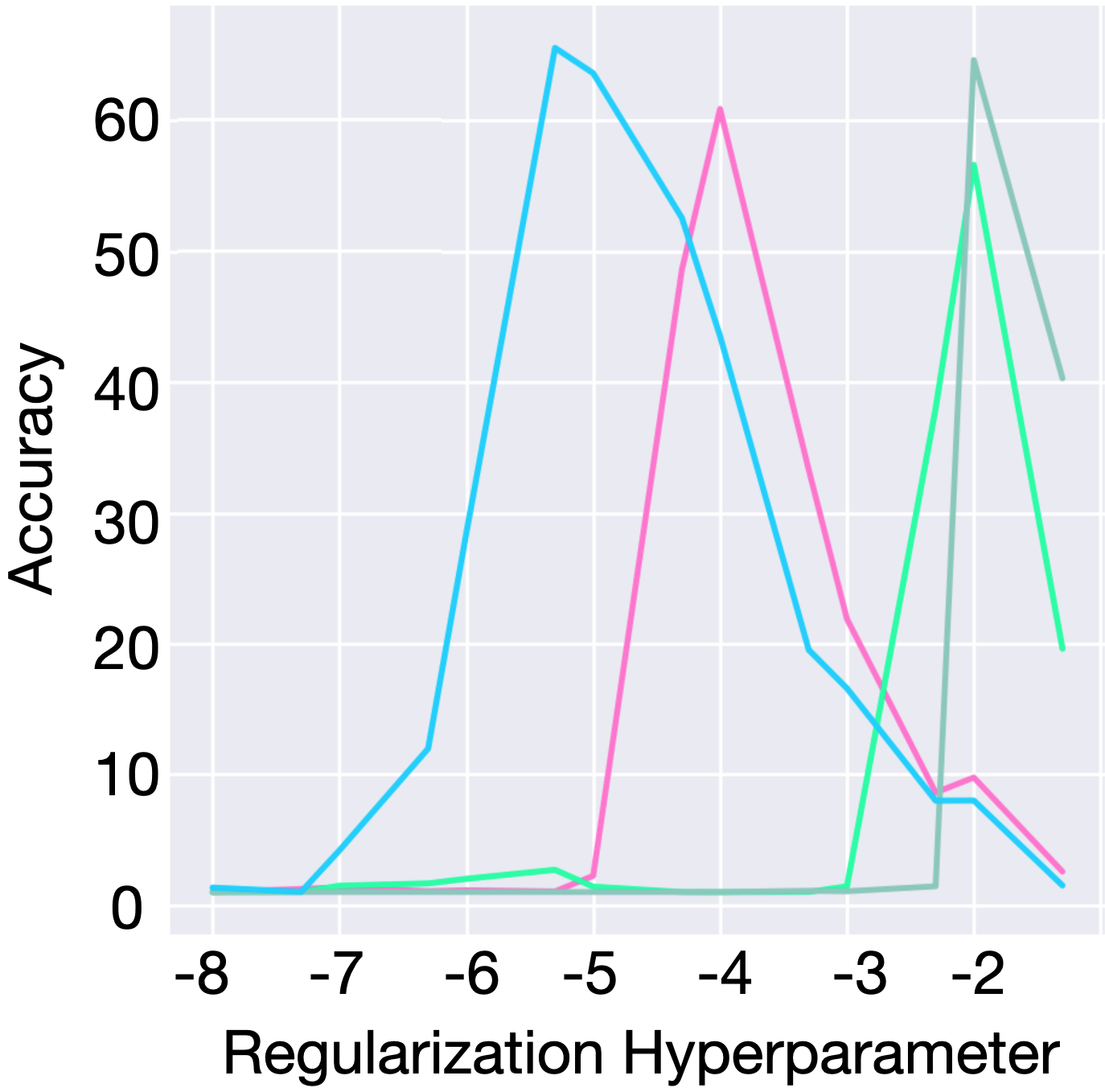}
        \caption{ResNet-50}
    \end{subfigure}
    \caption{Visualization of accuracy of VGG-like and ResNet-50 architectures at $0.95$ sparsity across different values of $\xi$ and $\psi$ on CIFAR-100.  On the x-axis we plot the regularization coefficients on a $\log_{10}$ scale. }
    \label{fig:c100_reg_coeff}
\end{figure}

We plot accuracy of one model run according to different values of $\xi$ and $\psi$. In our experiments, we set $\xi = \psi$ in order to eliminate the advantage of our approach having an extra hyperparameter over the $L_1$ penalty and MCP which does not have the hierarchical term.  In our main results, we report accuracy for each regularization approach based on the best performing regularization hyperparameters, although we note that for HALO, there are often a few hyperparameter choices which perform comparably or better than competing methods. Results for CIFAR-100 are given in Figure~\ref{fig:c100_reg_coeff}; results are similar for other datasets and networks. Results indicate that HALO has a wider range of ``acceptable`` parameter values while attaining higher accuracy. 

\subsection{Learning Different Types of Sparsity}
\label{sec:spars_type}

We have shown that our approach can prune network architectures in order to achieve small models with little drop in performance. We  further discuss the type of sparsity learned through the HALO penalty. In particular, we highlight that the HALO penalty performs monotonic penalization, learns structured sparsity, and learns low-rank feature representations.

\textbf{Approximately Structured Penalization}: To take advantage of unstructured sparsity, sparse libraries or special hardware is required to deploy such networks, and recent work has aimed at pruning layers, filters, or channels of the network \cite{alvarez2017compression, li2016pruning, wen2017coordinating, zhang2018learning}. We note that while it is not guaranteed for HALO to learn sparse group representations, our procedure learns to nearly sparsify complete layers yielding more efficient networks without needing sparse libraries or other mechanisms as shown in the main paper. We find that for early layers (before layer 5) the sparsity ratio is low, for middle layers (layers 5-8) there is a sharp increase in the sparsity level, and above layer 8, layers are near fully sparse.  Interestingly layer 2 exhibits a high amount of sparsity (approximately 70\%) on CIFAR-100, and the sparsity seems to exhibit a pattern every few layers. The pattern appears to arise from the structure of the VGG architectures which separate convolutional layers with max-pooling layers with large jumps or discontinuities occurring after max-pooling layers.

\textbf{Low-Rank Penalization}:  We additionally plot the normalized cumulative sum of the eigenvalues of the covariance matrix generated from the outputs of each convolutional layer, which reflects the dimensionality of the output feature space.  For the CIFAR-100 training set there is a sharp trend which plateaus at $1$ after only very few eigenvalues. This indicates that the covariance matrix of the outputs is low-rank, and that the HALO penalty learns a model producing a low-dimensional representation of the  feature spaces, much as other low-rank factorization approaches which might apply a low-rank matrix factorization (like PCA) to intermediate layers. Although untested, this type of representation is also learned to de-correlate and prune filters in \cite{Cuadros_2020_WACV}. We note that as with the structured penalization, the intermediate layers have the largest effective dimensionality. Along with the increase in sparsity during the intermediate layers, this may imply that the middle layers are the most feature-rich for image classification, and are an important subject for future work.

\begin{figure}[ht]
    \centering
    \begin{subfigure}[b]{0.15\textwidth}
        \includegraphics[width=\textwidth]{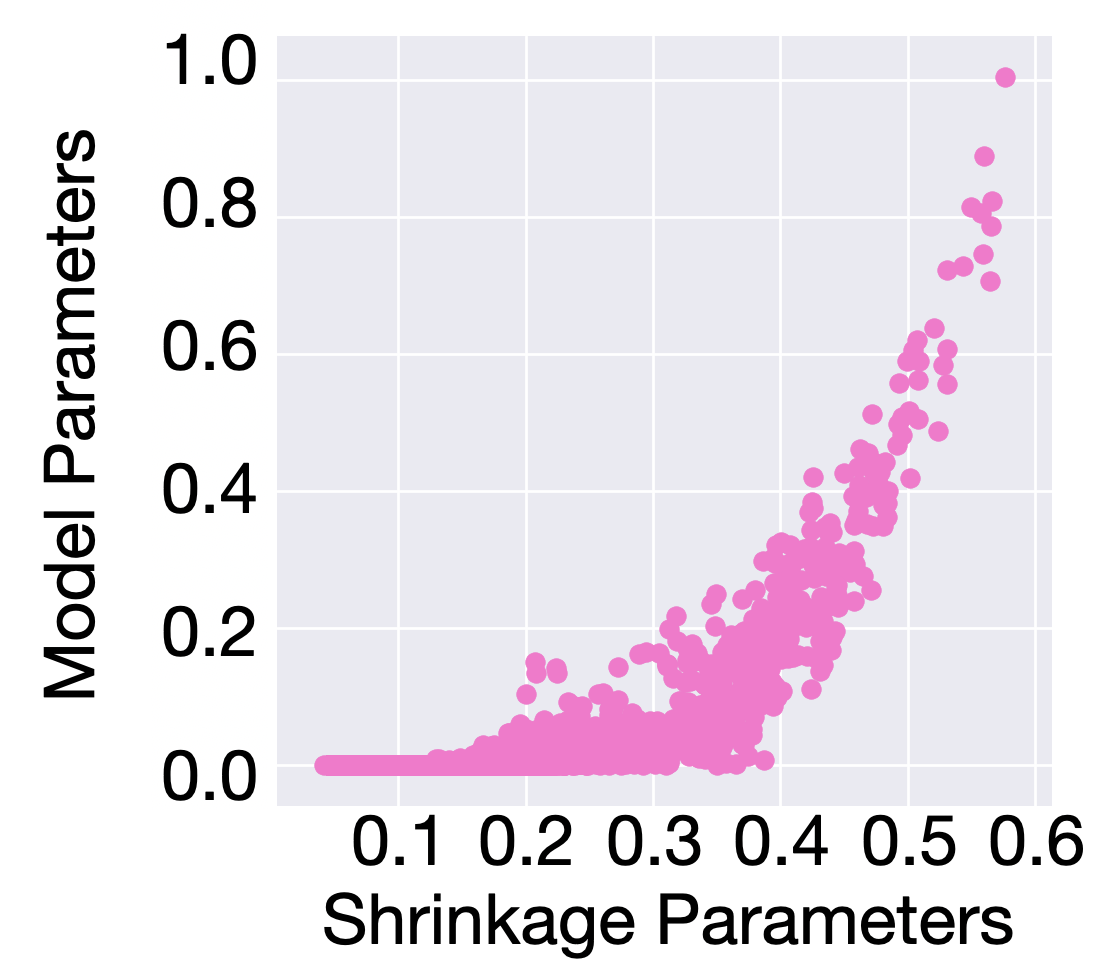}
        \caption{Monotonic}
    \end{subfigure}
    \begin{subfigure}[b]{0.14\textwidth}
        \includegraphics[width=\textwidth]{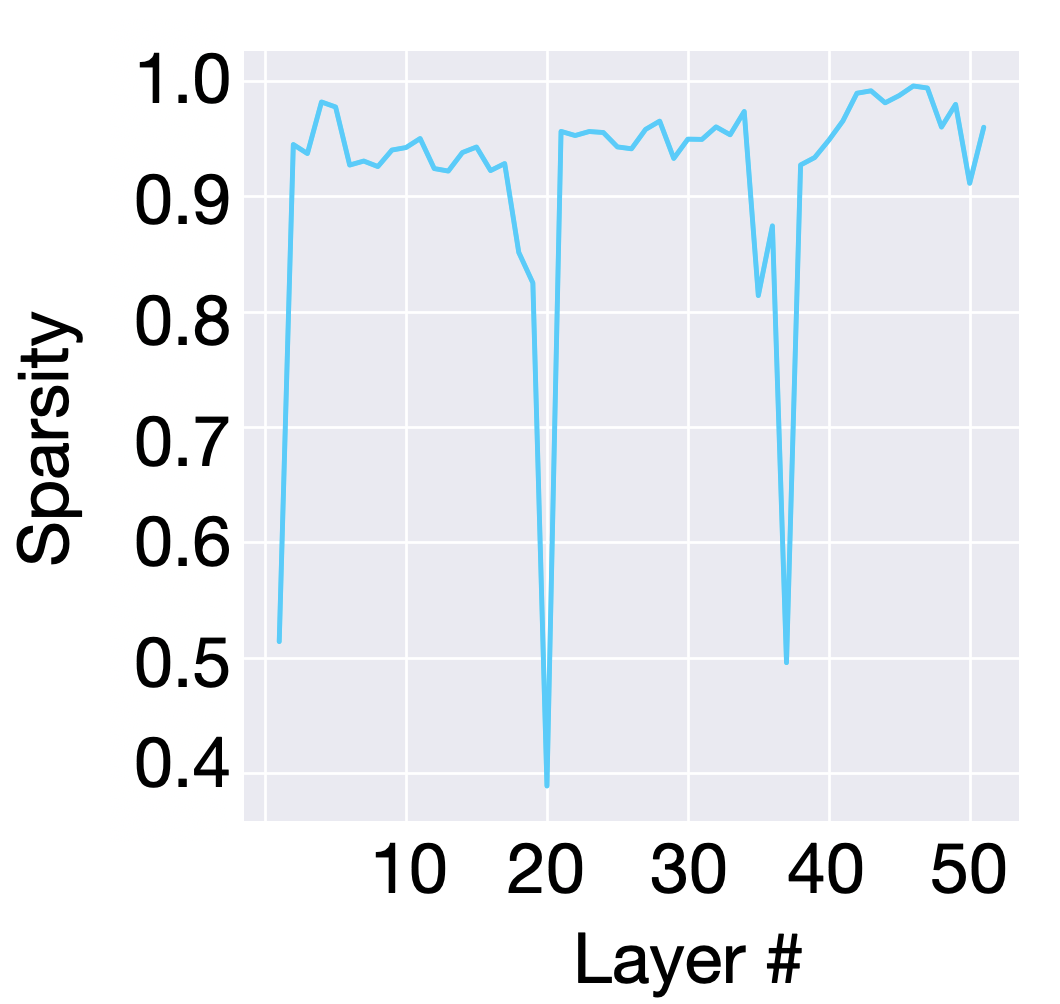}
        \caption{Structured}
    \end{subfigure}
    \begin{subfigure}[b]{0.15\textwidth}
        \includegraphics[width=\textwidth]{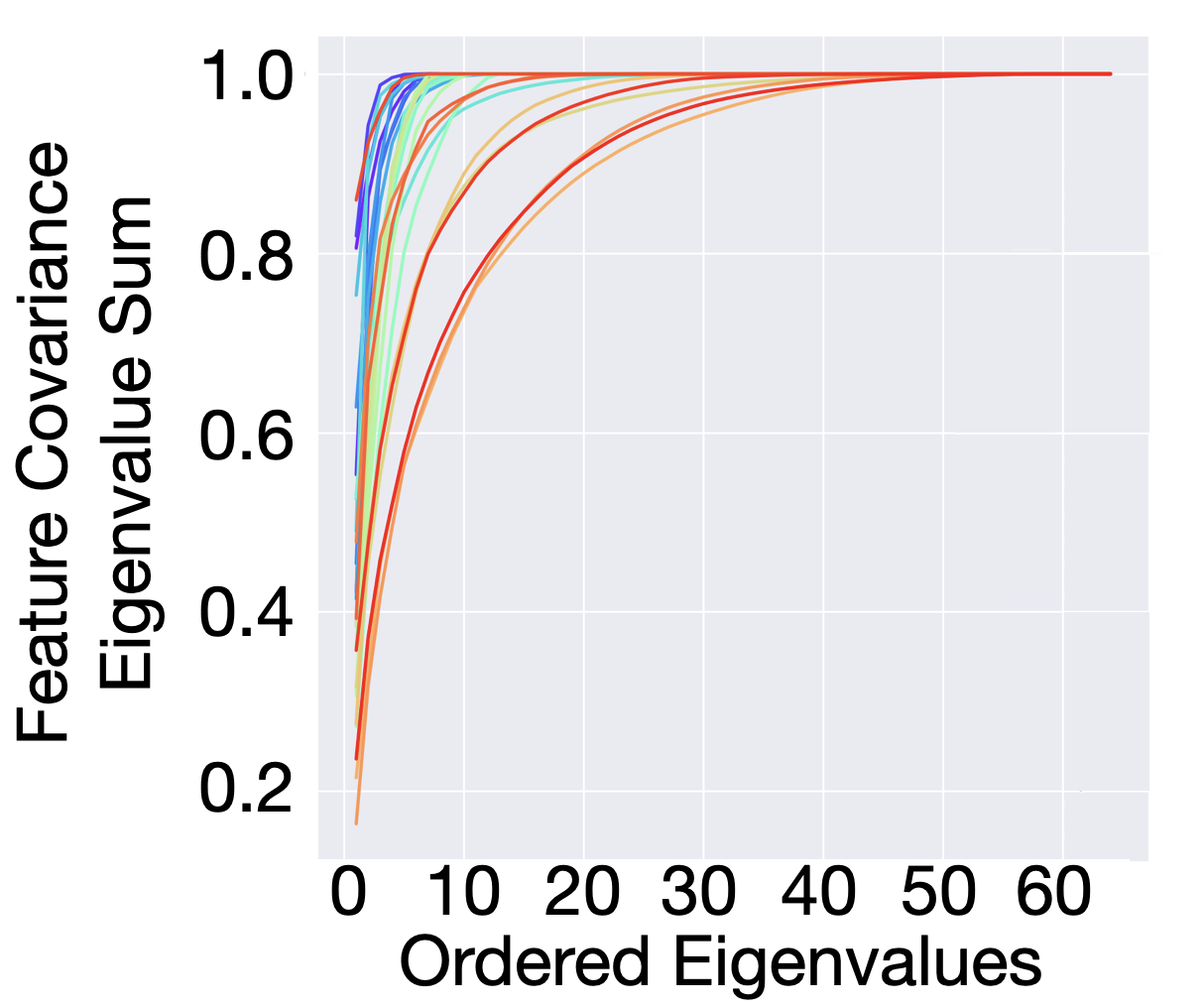}
        \caption{Low-rank}
    \end{subfigure}
    \caption{Different types of penalties learned by HALO for ResNet-50 on CIFAR-100 at $0.95$ sparsity. (a) A random sample of 10,000 regulrization coefficients and weight parameters illustrating a monotonically increasing trend. (b) Sparsity by layer for CIFAR-10/0 indicating entire layers are sparse (c) Normalized cumulative sum of eigenvalues of the output covariance matrix indicating a small number of important features. Colors ranging from violet to red indicate layer number from low to high.}
    \label{fig:res50_learn_sparsity}
\end{figure}

Additional sparsity plots for ResNet-50 on CIFAR-100 are provided in Figure~\ref{fig:res50_learn_sparsity}. A key difference with those of VGG is that the large drops in sparsity per layer are from the downsample layers at the end of each block in ResNet which lower the width and height and increase the number of channels.

\begin{figure}[ht]
    \centering
    \begin{subfigure}[b]{0.15\textwidth}
        \includegraphics[width=\textwidth]{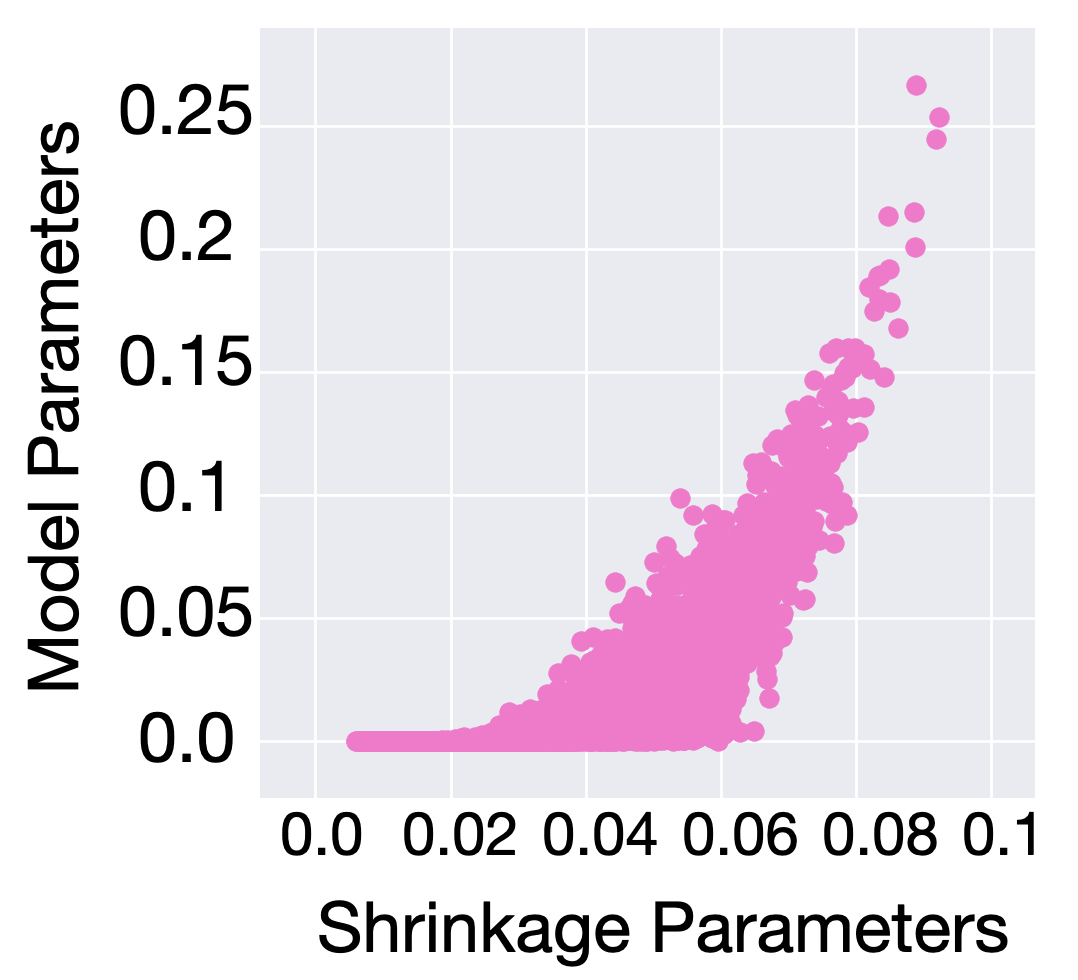}
        \caption{Monotonic}
    \end{subfigure}
    \begin{subfigure}[b]{0.145\textwidth}
        \includegraphics[width=\textwidth]{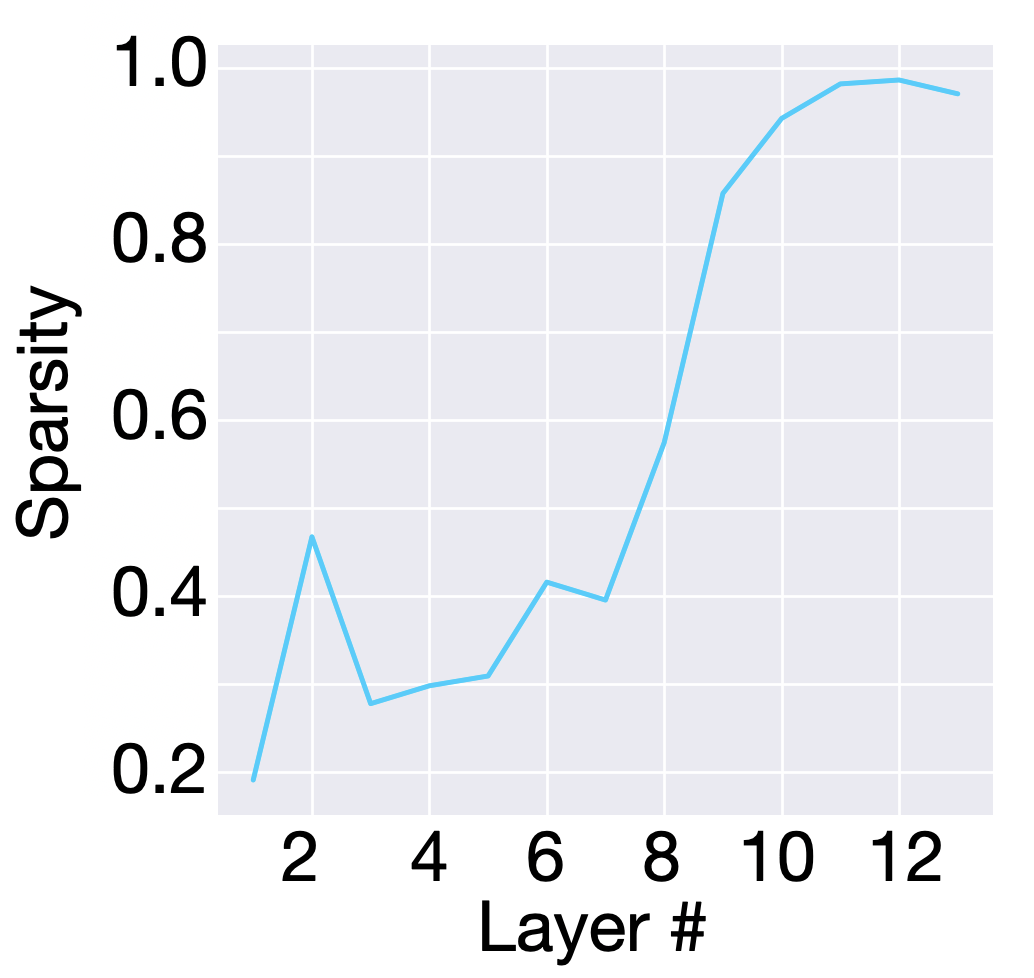}
        \caption{Structured}
    \end{subfigure}
    \begin{subfigure}[b]{0.16\textwidth}
        \includegraphics[width=\textwidth]{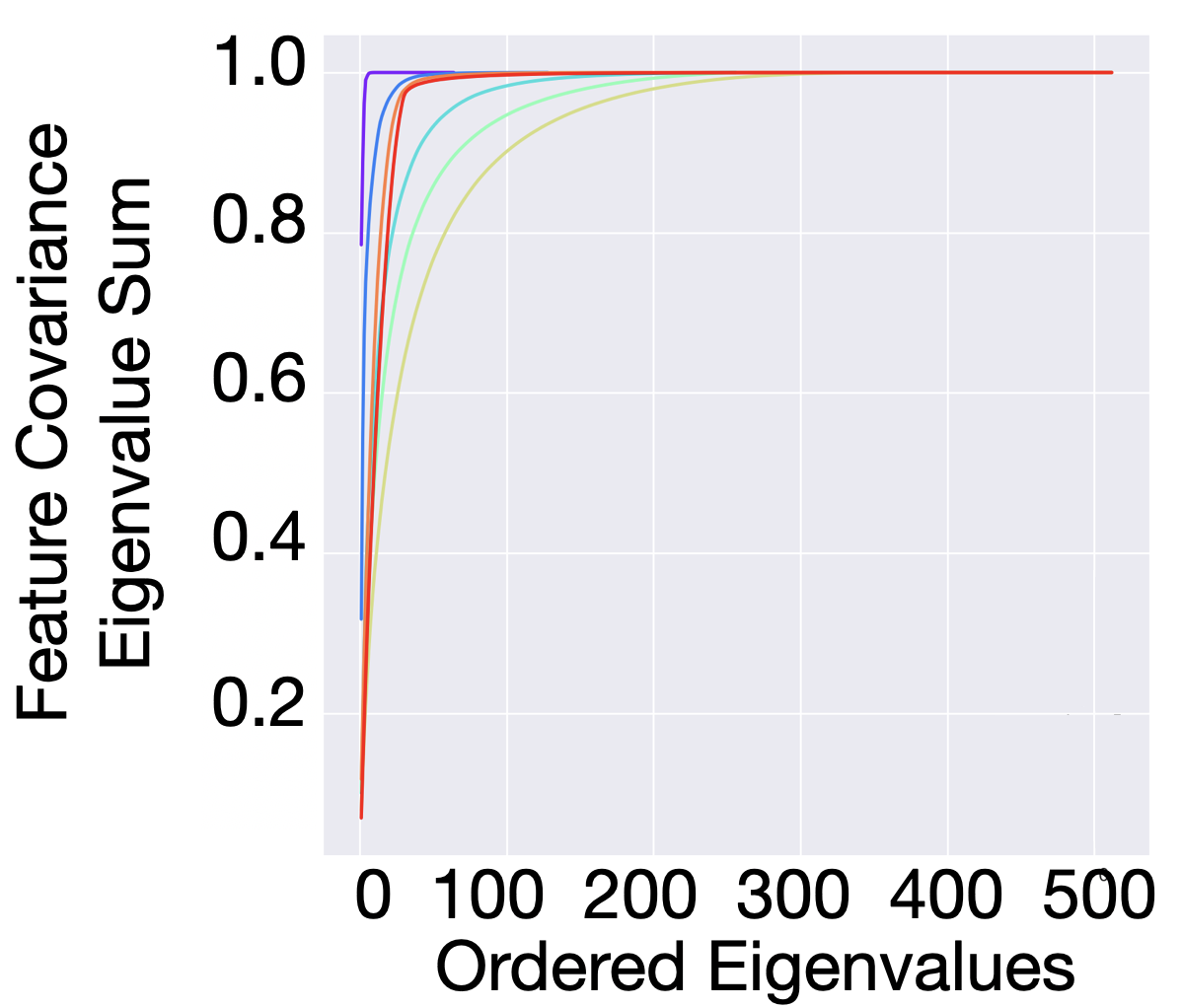}
        \caption{Low-rank}
    \end{subfigure}
    \caption{Different types of penalties learned by HALO for VGG-like on CIFAR-100 at $0.85$ sparsity.}
    \label{fig:vgg16_learn_sparsity_low}
\end{figure}

\begin{figure}[ht]
    \centering
    \begin{subfigure}[b]{0.145\textwidth}
        \includegraphics[width=\textwidth]{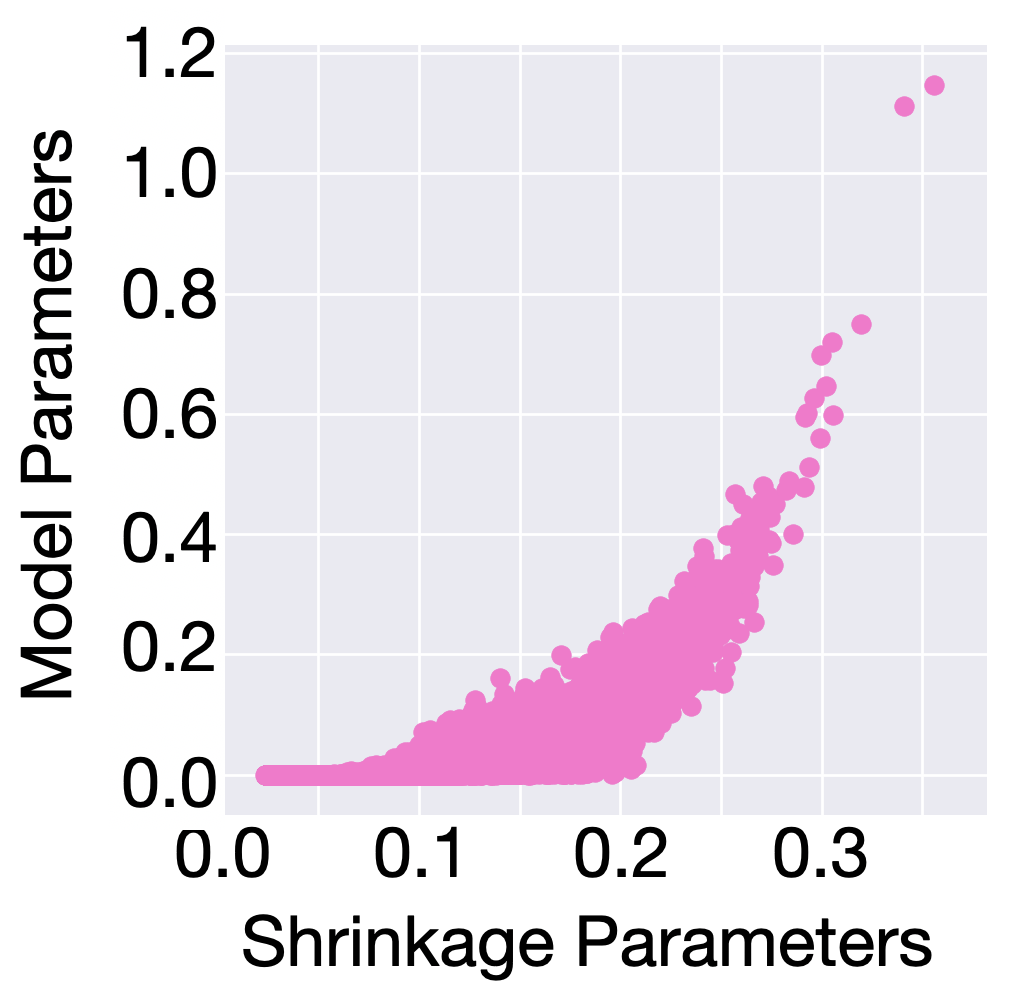}
        \caption{Monotonic}
    \end{subfigure}
    \begin{subfigure}[b]{0.145\textwidth}
        \includegraphics[width=\textwidth]{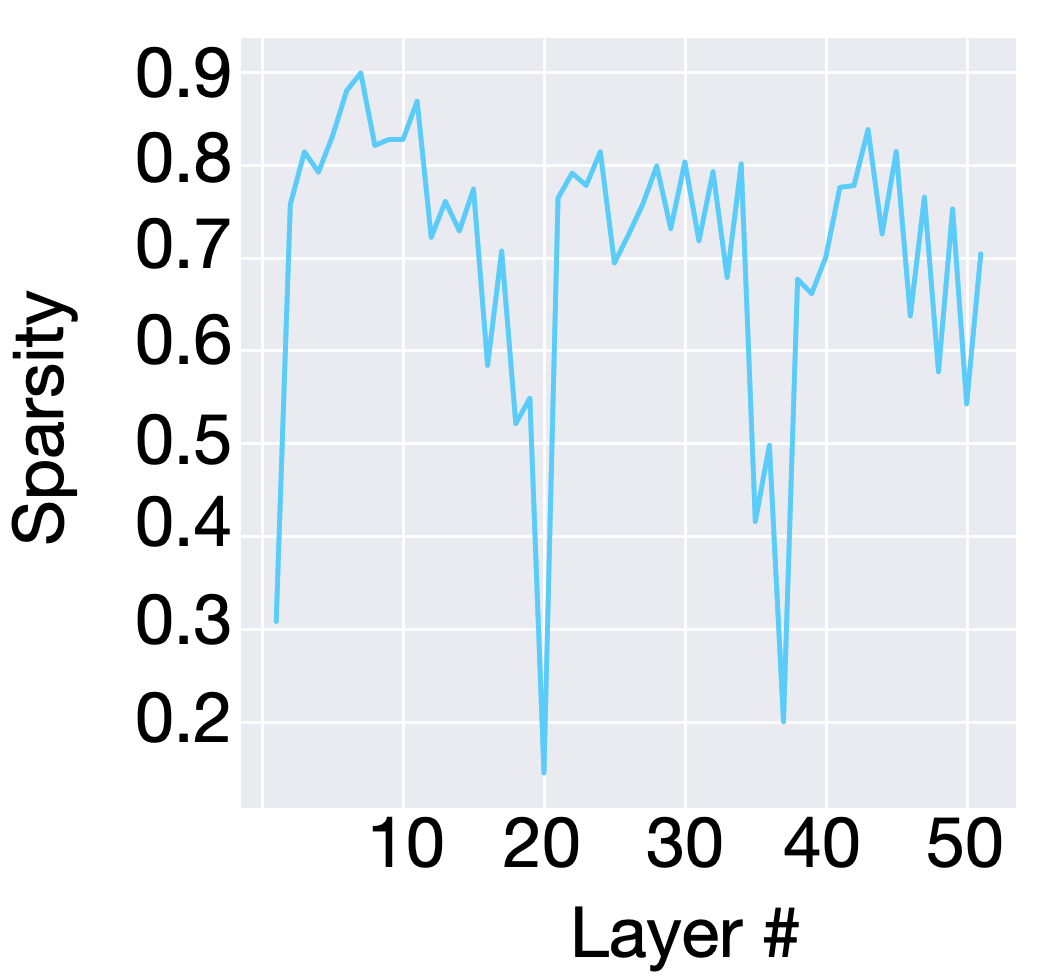}
        \caption{Structured}
    \end{subfigure}
    \begin{subfigure}[b]{0.16\textwidth}
        \includegraphics[width=\textwidth]{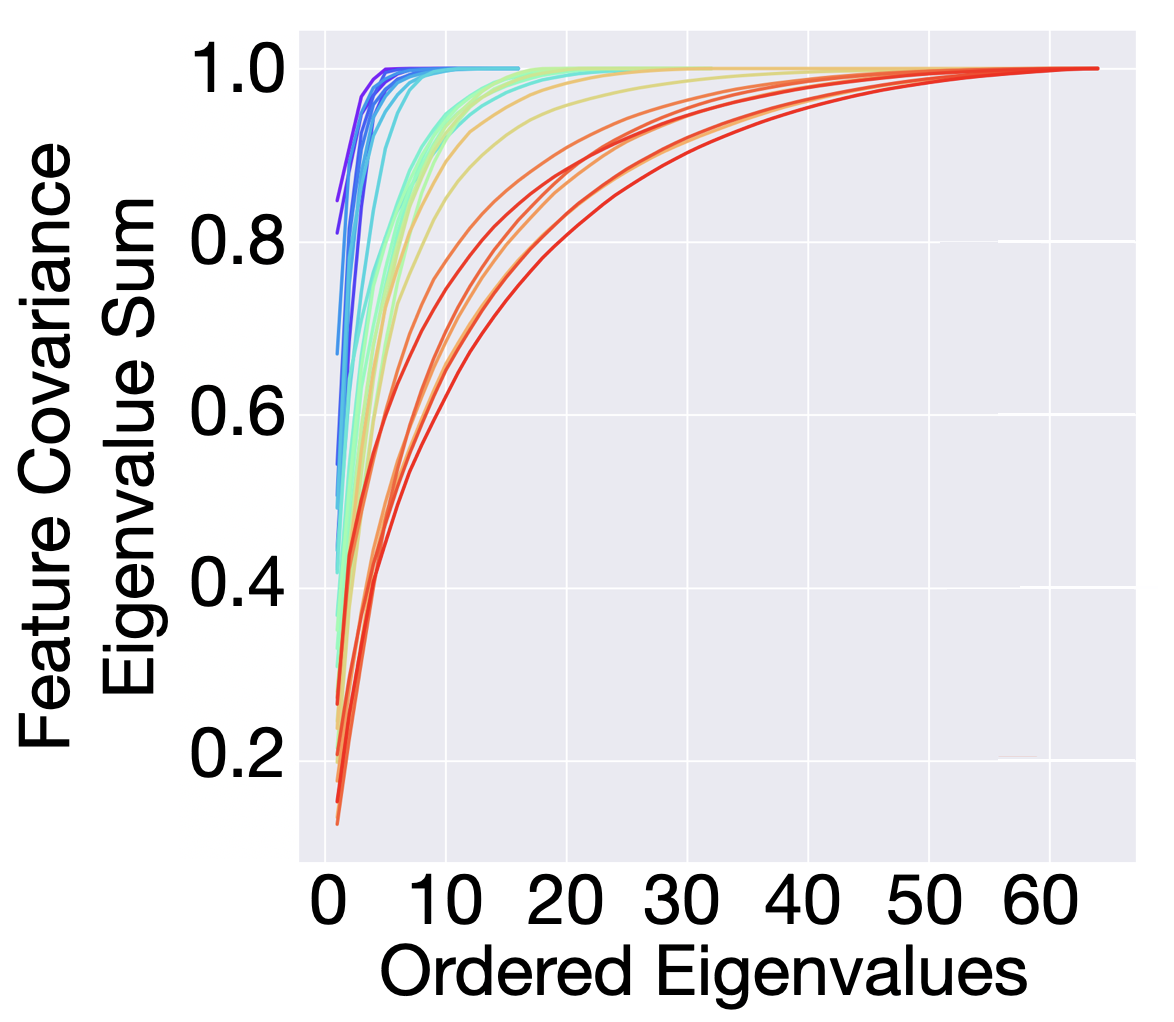}
        \caption{Low-rank}
    \end{subfigure}
    \caption{Different types of penalties learned by HALO for ResNet-50 on CIFAR-100 at $0.7$ sparsity.}
    \label{fig:res50_learn_sparsity_low}
\end{figure}

We additionally provide sparsity plots for VGG and ResNet-50 when the sparsity is below $0.95$ and the accuracy is maintained. The characteristics of sparsity are similar to those at the higher 0.95 sparsity level, which means that these are properties achievable with minimal drop in performance as seen in Figures~\ref{fig:vgg16_learn_sparsity_low} and \ref{fig:res50_learn_sparsity_low}.

\subsection{Training Time}
\label{sec:rt}

We summarize training time reported as the number of seconds taken to train a VGG-like network on CIFAR-100 with each penalized training approach in Table~\ref{tab:time}. We find that regularization approaches are slower for a single epoch, however pruning methods require two or more stages of training and in our experiments require double the number of epochs for training. 

\begin{table}[ht]
    \centering
    \begin{tabular}{|c|c|}
        \hline
        Method & Train Time \\\hline
        Baseline & \textbf{19.07} $(\pm 0.24)$ \\
        $L_1$ & 23.98 $(\pm 0.54)$ \\
        SWS & 25.90 $(\pm 0.37)$\\
        HALO & 31.63 $(\pm 0.23)$\\\hline
    \end{tabular}
    \caption{Average training time for a single epoch using a penalty over 5 runs with standard deviation.}
    \label{tab:time}
\end{table}

\subsection{Sparsity During Training}

\begin{figure}[ht]
    \centering
    \begin{subfigure}[b]{0.25\textwidth}
        \includegraphics[width=\textwidth]{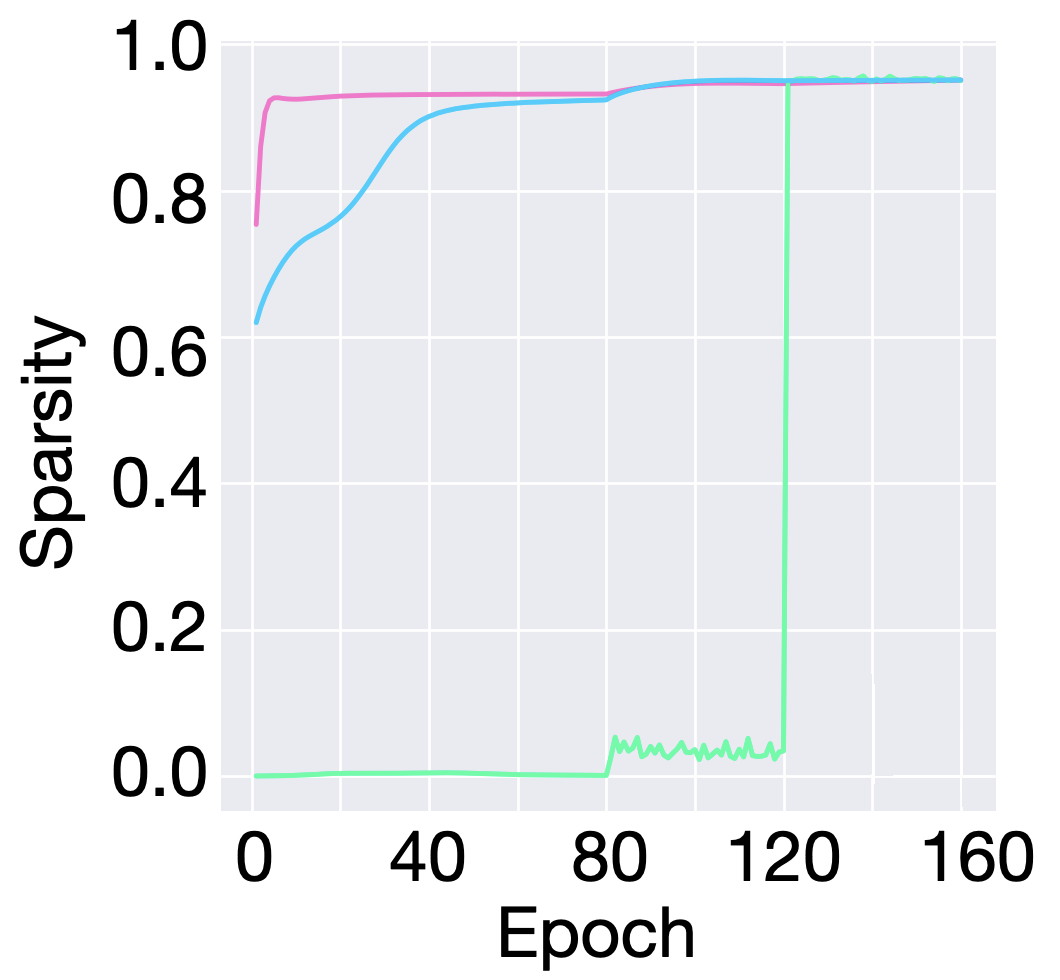}
        \caption{VGG-like}
    \end{subfigure}
    \begin{subfigure}[b]{0.25\textwidth}
        \includegraphics[width=\textwidth]{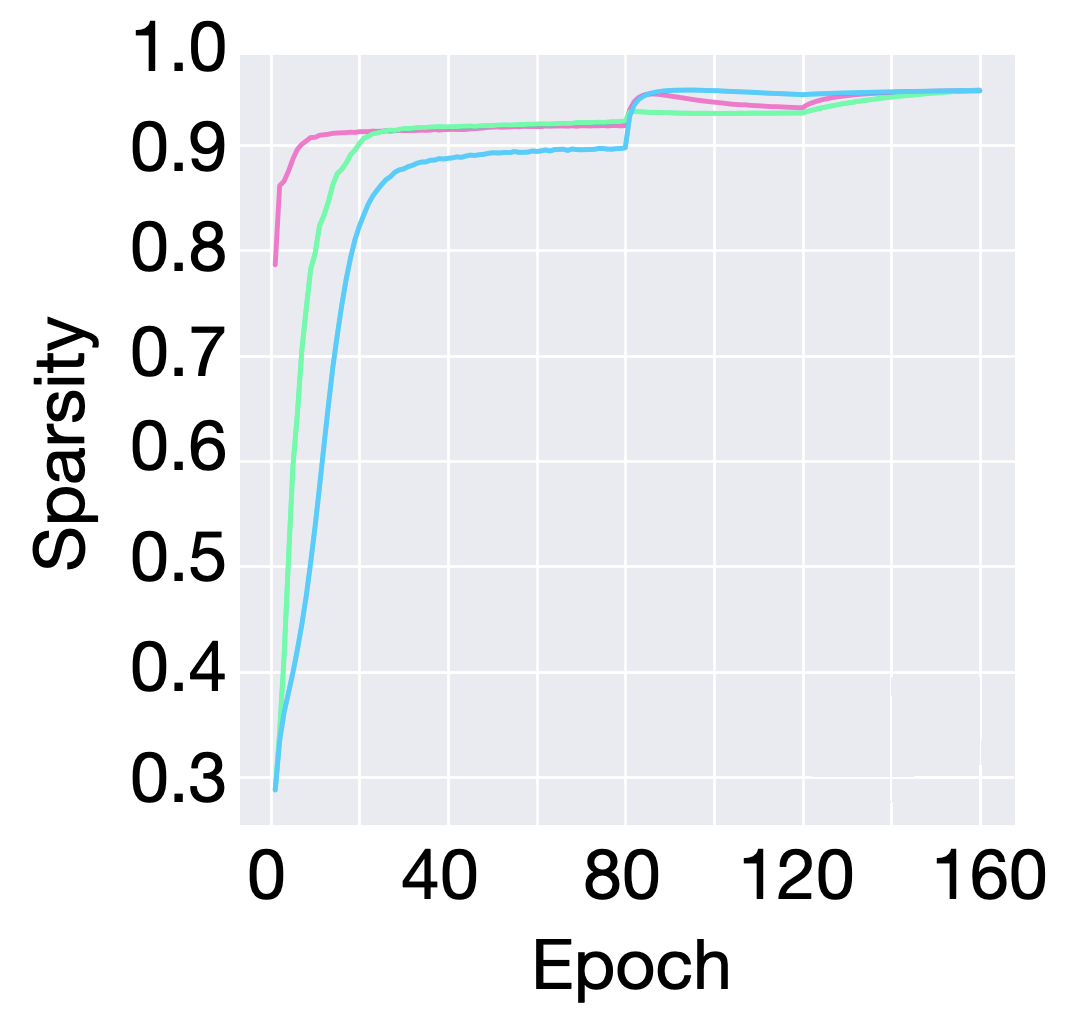}
        \caption{ResNet-50}
    \end{subfigure}
    \caption{Amount of sparsity during training on CIFAR-100. Pink: $L_1$, Green: SWS, Blue: HALO}
    \label{fig:c100_sparsity_threshold}
\end{figure}

We plot the amount of sparsity in the model during training for both the VGG-like architecture and ResNet architecture (Figure~\ref{fig:c100_sparsity_threshold}) by counting the number of parameters in the model which are smaller in magnitude than the $95$th percentile weight from a fully-trained network trained with the same penalty.  For both architectures, we find that adaptive penalties pushes coefficients to zero at a slower rate than the non-adaptive penalty. The threshold for the weights trained with adaptive penalty in both architectures is also smaller than the threshold for the $L_1$ penalty as seen in the lower sparsity at initial epoch indicating that the weights are further shrunk to $0$ over the $L_1$ penalty. 

In contrast to the $L_1$ and HALO penalties, on VGG, the SWS penalty penalizes relatively late in the training phase and attains a small threshold, several orders of magnitude smaller than the $L_1$ and HALO penalties indicating it has set a majority of the weights to nearly zero at the end of training. 

\subsection{Sparsity Overlap}
\label{sec:so}

\begin{figure}[ht]
    \centering
    \begin{subfigure}[b]{0.25\textwidth}
        \includegraphics[width=\textwidth]{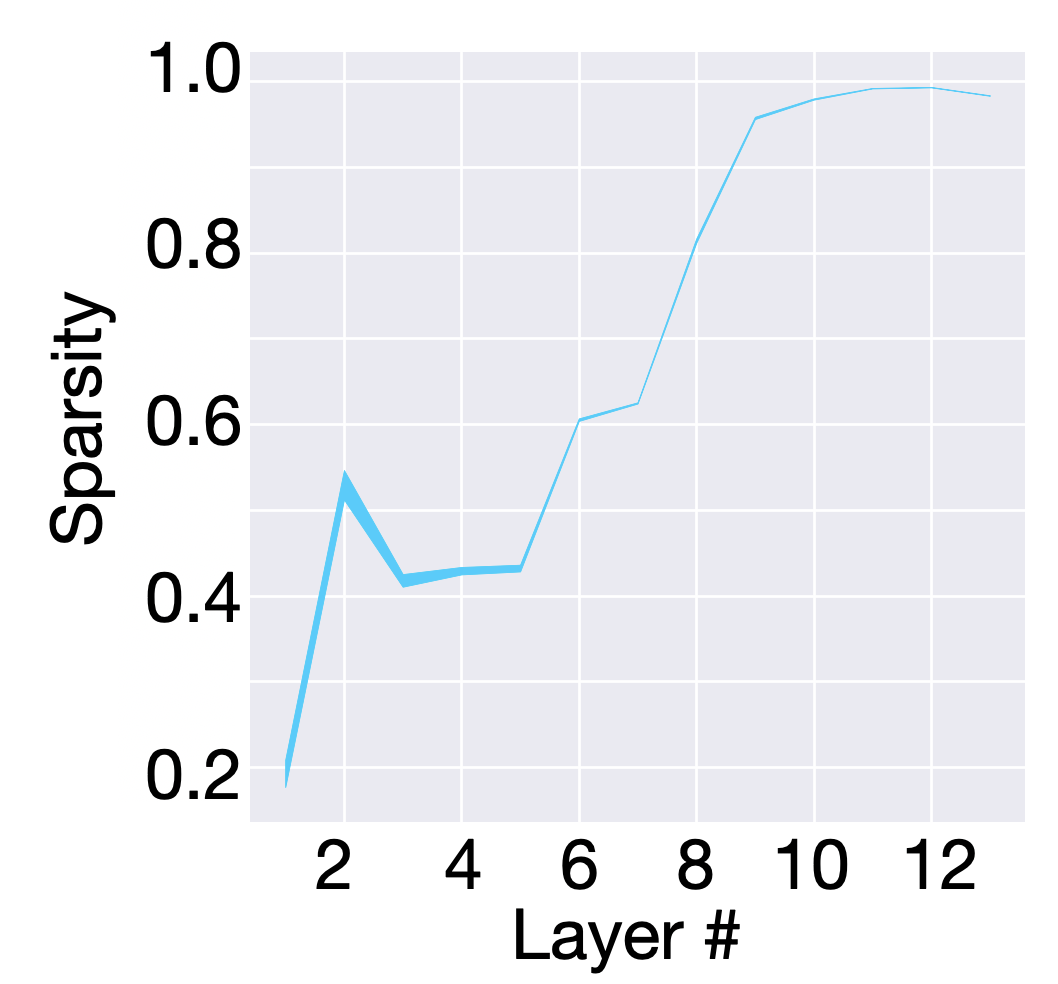}
        \caption{VGG on C100}
    \end{subfigure}
    \begin{subfigure}[b]{0.25\textwidth}
        \includegraphics[width=\textwidth]{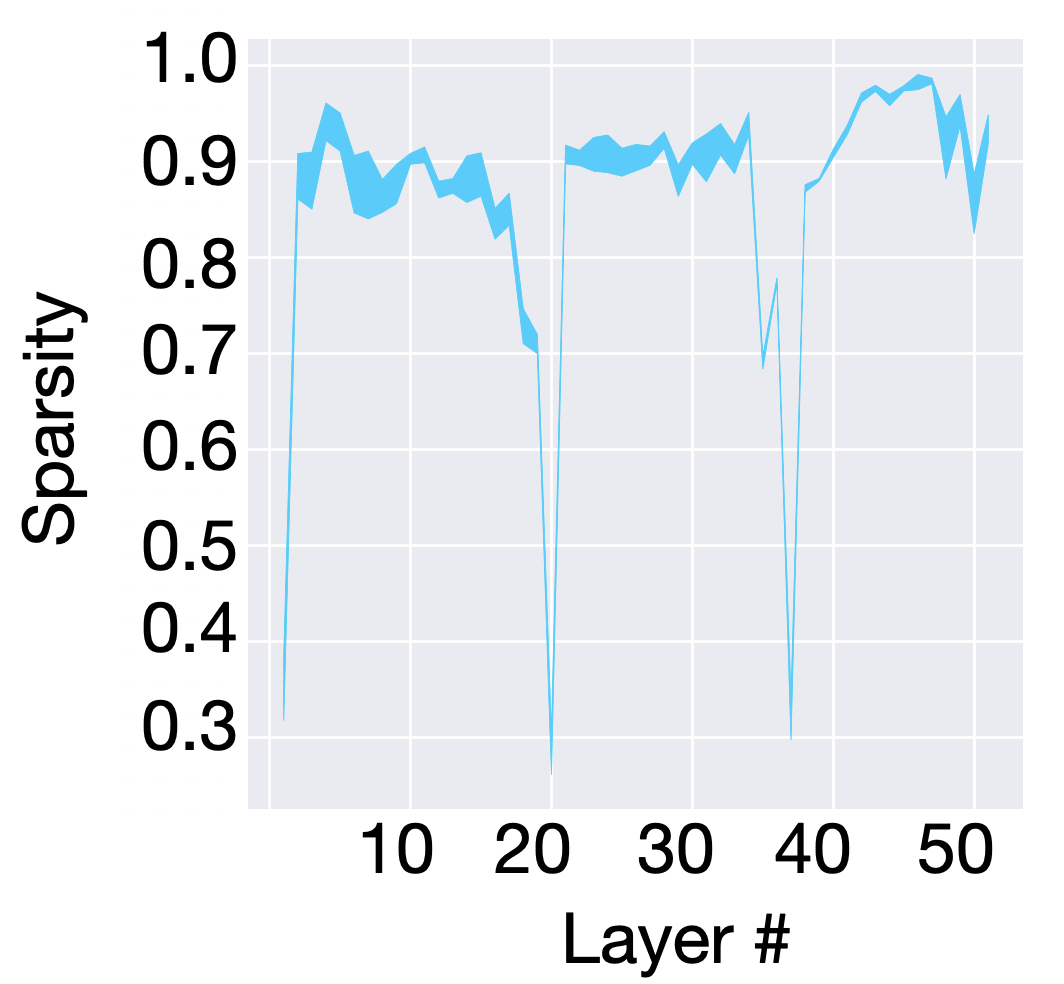}
        \caption{Res50 on C100}
    \end{subfigure}
    \caption{Sparsity overlap ($SO$) averaged over four runs for different HALO models compared with one another.}
    \label{fig:overlap}
\end{figure}

We investigate how similar the learned parameters are to one another over multiple runs of HALO and summarizes the results in Figure~\ref{fig:overlap}. The sparsity overlap ($SO$) is computed as the Jacard similarity of zero weights in two models; that is, let $\mathcal{A}$ be the set of zero weights for model $A$ and $B$ be the set of weights for model $\mathcal{B}$. Then the sparsity overlap for model $A$ and $B$ is defined as $SO = \frac{\mathcal{A} \cap \mathcal{B}}{\mathcal{A} \cup \mathcal{B}}$.

The shape of the sparsity overlap in Figure~\ref{fig:overlap} follows that of the structured penalization, which means that the higher the level of sparsity in a layer the higher the percentage overlap of sparsity across multiple runs of HALO. This is straightforward for the highly sparse layers. For the less sparse layers, this means that there are multiple winning tickets, which can be expected due the large number of weights.

\end{document}